# Challenges, Advances, and Evaluation Metrics in Medical Image Enhancement: A Systematic Literature Review

Chun Wai Chin[1] Haniza Yazid[2*] Hoi Leong Lee[3*]
[1,2,3]Faculty of Electronic Engineering & Technology (FKTEN), Universiti Malaysia Perlis (UniMAP), Ulu Pauh Campus, 02600 Arau, Perlis.
[1]cchin535@gmail.com [2*]hanizayazid@unimap.edu.my
[3*]hoileong@unimap.edu.my

## Abstract

Medical image enhancement is crucial for improving the quality and interpretability of diagnostic images, ultimately supporting early detection, accurate diagnosis, and effective treatment planning. Despite advancements in imaging technologies such as X-ray, CT, MRI, and ultrasound, medical images often suffer from challenges like noise, artifacts, and low contrast, which limit their diagnostic potential. Addressing these challenges requires robust preprocessing, denoising algorithms, and advanced enhancement methods, with deep learning techniques playing an increasingly significant role. This systematic literature review, following the PRISMA approach, investigates the key challenges, recent advancements, and evaluation metrics in medical image enhancement. By analyzing findings from 39 peer-reviewed studies, this review provides insights into the effectiveness of various enhancement methods across different imaging modalities and the importance of evaluation metrics in assessing their impact. Key issues like low contrast and noise are identified as the most frequent, with MRI and multi-modal imaging receiving the most attention, while specialized modalities such as histopathology, endoscopy, and bone scintigraphy remain underexplored. Out of the 39 studies, 29 utilize conventional mathematical methods, 9 focus on deep learning techniques, and 1 explores a hybrid approach. In terms of image quality assessment, 18 studies employ both reference-based and non-reference-based metrics, 9 rely solely on reference-based metrics, and 12 use only non-reference-based metrics, with a total of 65 IQA metrics introduced, predominantly non-reference-based. This review highlights current limitations, research gaps, and potential future directions for advancing medical image enhancement.

## 1.0 Introduction

### 1.1 Overview

Medical imaging has revolutionized modern healthcare by enabling non-invasive visualization of the human body's internal structures and functions. Advanced imaging modalities such as X-ray, computed tomography (CT), magnetic resonance imaging (MRI), and ultrasound are indispensable tools for diagnosing a wide range of medical conditions.

However, the utility of these imaging techniques is often compromised by inherent challenges, including noise, artifacts, and low contrast, which obscure critical diagnostic details. Enhancing the quality of medical images is, therefore, a fundamental step in ensuring accurate diagnoses and optimal treatment outcomes.

Medical image enhancement encompasses a broad spectrum of preprocessing techniques designed to improve the visual quality of images. These techniques range from traditional methods such as histogram equalization and Gaussian filtering to sophisticated approaches involving deep learning models. Despite significant advancements, medical image enhancement faces several challenges, such as balancing noise reduction with detail preservation and mitigating artifacts without introducing unnatural distortions. Furthermore, the effectiveness of enhancement methods varies across imaging modalities and clinical applications, necessitating the development of modality-specific solutions.

### 1.2 Motivation

The motivation for conducting this systematic literature review (SLR) arises from the lack of comprehensive reviews focusing on recent enhancement methods that address various image quality issues across multiple medical imaging modalities. While some studies emphasize image restoration or resolution, they often neglect enhancement techniques tailored to specific imaging modalities [1], [2]. Other reviews primarily concentrate on recent advancements in denoising algorithms, addressing only one aspect of image quality issues, which limits their comprehensiveness [3], [4]. Similarly, some reviews focus on specific enhancement algorithms, such as those aimed at improving image resolution, or on particular areas like 3D medical image processing and image fusion [5], [6], [7]. Furthermore, there are reviews restricted to particular imaging modalities, such as MRI, rather than encompassing the diversity of medical imaging techniques [8].

This SLR aims to bridge these gaps by comprehensively analysing the challenges, advancements, and evaluation metrics in medical image enhancement. By synthesizing findings from 39 research studies, this review examines the strengths and limitations of various enhancement techniques, evaluates their performance using standardized metrics, and highlights emerging trends in the field. The findings of this review will serve as a valuable resource for researchers and practitioners seeking to advance the state of the art in medical image enhancement, ultimately contributing to improved diagnostic accuracy and patient care.

### 1.3 Objectives

The review consists of three main objectives, which are as follows:

a) To identify image quality issues in modern medical imaging modalities.

b) To analyse traditional, deep learning-based, and hybrid approaches in medical imaging for their effectiveness in improving image quality and diagnostic accuracy.
c) To investigate commonly used and new quantitative metrics for assessing enhancement methods in medical imaging.

### 1.4 Key Features / Contributions of The Review

The review emphasizes several key aspects of the studies, which are outlined as follows:

a) Identification of Image Quality Issues and Their Correlation with Modalities and Datasets
This review identifies and analyses prevalent image quality issues across major medical imaging modalities such as X-ray, CT, MRI, and ultrasound. It further correlates these challenges with associated datasets, providing dataset links to enhance reproducibility and future research efforts.
b) Comprehensive Analysis of Enhancement Techniques
An in-depth evaluation of both traditional and deep learning-based medical image enhancement methods is provided, detailing their principles, applications, strengths, and limitations.
c) Insights into Evaluation Metrics
The review examines key evaluation metrics for assessing image quality, contrast, and denoising algorithm, offering guidance on their selection and application for different enhancement tasks.

## 2.0 Systematic Survey Methodology

This review paper utilizes Systematic Reviews and Meta-Analyses (PRISMA) guidelines [9] to investigate relevant studies on the selected topic.

### 2.1 Research Questions

This review outlines several key questions in different aspects to guide researchers in the future development of effective medical image enhancement algorithms:

1. **Challenges with Image Quality in Modern Medical Imaging Modalities**
   a) What are the common issues related to image quality (e.g., noise, artifacts, low contrast) in recent imaging modalities?
2. **Advances and Comparisons in Image Enhancement Techniques**
   a) What are the most widely applied image enhancement techniques (e.g., deep learning-based, traditional image processing, hybrid approaches) in modern medical imaging?
   b) How do these methods compare in improving image quality, contrast, resolution, and overall diagnostic accuracy across different imaging modalities?
   c) What are the strengths and limitations of existing comparative studies that benchmark these methods across multiple medical imaging modalities?
3. **Evaluation Metrics for Image Enhancement Techniques**

a) What are the commonly used quantitative metrics (reference-based and non-reference-based) for evaluating the effectiveness of image enhancement techniques in medical imaging studies?
b) What is the indication in term of image quality for each Image Quality Assessment (IQA) metrics?

### 2.2 Search Strategy

A systematic approach was used to identify relevant literature for the review. Article searches were conducted through multiple electronic databases to ensure comprehensive coverage. The search was restricted to studies published within the last five years to capture the most recent advancements in the field. The databases included Science Direct and Web of Science (WoS). The search formula for each of the three databases was as follows: ("medical image enhancement" AND (“contrast " OR "noise" OR "uneven background”)). Boolean operators “AND” or “OR” were used in searching the papers. The search was limited to only complete English textual articles and included research articles only.

### 2.3 Eligibility Criteria

#### A. Inclusion Criteria

Studies were deemed eligible if they met the following requirements:

a) Included the selected search keywords in abstract and/or title and/or keywords of the study.
b) Articles focusing on image enhancement techniques applied to medical imaging (e.g., MRI, CT, X-rays, ultrasound, histopathology slides).
c) Studies involving medical imaging datasets for diagnostic or research purposes.
d) Selection is limited to studies published in the last 5 years, from 2020 – 2025
e) Research that explicitly addresses methods to improve image quality, contrast, noise, blurring and colour imbalance in a clinical or diagnostic applications.
f) Studies that involved any enhancement techniques such as deep learning, traditional image processing, or hybrid approaches.
g) Full text English studies only.
h) Studies with quantitative or qualitative evaluation of image enhancement methods.

#### B. Exclusion Criteria

The following were excluded from the study:

a) Studies that were not able to be accessed.
b) Books, proceeding papers, letters, poster, short papers, survey or literature review and case reports.
c) Abstracts without full-text availability.
d) Studies focusing solely on non-medical applications of image enhancement.

e) Articles without empirical validation or results (e.g., purely theoretical works).
f) Exclusion of studies that do not involve human or clinical data (e.g., animal models without validation on clinical datasets).
g) Studies that were not able to provide details of the methodology.
h) Only partially IQA results were disclosed.
i) IQA was not performed on the developed enhancement method but with segmentation or classification results.
j) Paper that consists of super resolution and image fusion.

**2.4 Data Extraction**

The relevant data extracted were authors, publication year, image quality issues, types of medical images, datasets, details of the enhancement methods, software used, evaluation metrics for Image Quality Assessment (IQA), outcomes and its advantages and disadvantages.

**2.5 Quality Assessment Criteria**

To evaluate the quality of retrieved articles, a standardized and systematic approach was employed to evaluate the quality and credibility of the selected articles. Two independent reviewers conducted the assessment, ensuring consistency and minimizing bias. The evaluation process was guided by questions adapted from existing frameworks [10] and customized to align with the focus on medical image enhancement. Some questions were excluded or revised to better reflect the scope of this review, which centres on image enhancement methods.

Each question was assigned a score: “2” if the criterion was fully met, “1” if partially met or lacked sufficient detail, and “0” if not addressed. For criteria that were not applicable, “NA” was recorded. This scoring system enabled a structured and objective review process. The quality assessment questions are as follows:

1. Is the study objective clearly stated and relevant to medical image enhancement?
2. Does the study outline a robust and detailed research design?
3. Are the characteristics of the datasets or imaging modalities explicitly described?
4. Are the image enhancement methods clearly defined and adequately detailed?
5. Does the study focus on enhancing image quality, contrast, or denoising in a medical imaging application?
6. Are the evaluation metrics used to assess image enhancement techniques clearly defined and justified?
7. Does the study apply appropriate statistical or computational methods, and are they validated or verified?
8. Are the results and outcomes presented clearly and comprehensively?

9. Does the study acknowledge its limitations and discuss their implications?
10. Is there a well-supported and coherent conclusion that aligns with the study objectives?

## 3.0 Results

This section will primarily concentrate on presenting the search results obtained after implementing the survey methodology outlined in the previous section. Analyzing the quality of the data extracted from the reviewed articles also will be performed. Finally, it will highlight the current challenges in medical imaging modalities that impact image quality, explore advancements in contemporary medical image enhancement techniques, and examine the evaluation methods used to assess image quality.

## 3.1 Primary Search Results

The process of screening and narrowing down articles for analysis in this review was conducted systematically and was last updated on 24$^{th}$ December 2024 at 11:04 AM (Malaysia Time). As summarized in Figure 3.1, the initial search identified 326 records from two electronic databases: ScienceDirect (263 articles) and Web of Science (63 articles). Of these, 196 records were excluded based on accessibility issues and other criteria, such as the exclusion of books, conference proceedings, and studies published before 2020. This refinement resulted in 130 unique articles, which were further screened to remove duplicates and exclude papers based on their titles and abstracts that did not meet the inclusion criteria. During this phase, a total of 59 papers were removed. Following this, 71 articles underwent full-text eligibility assessment. Studies that did not align with the focus of the reviews such as those involving non-medical image datasets, video datasets, or animal tissue datasets were excluded. Additionally, studies lacking Image Quality Assessment (IQA) analysis, disclosing partial IQA results, or presenting non-absolute metrics were removed. After applying these rigorous criteria, 39 articles remained for detailed analysis in the review.

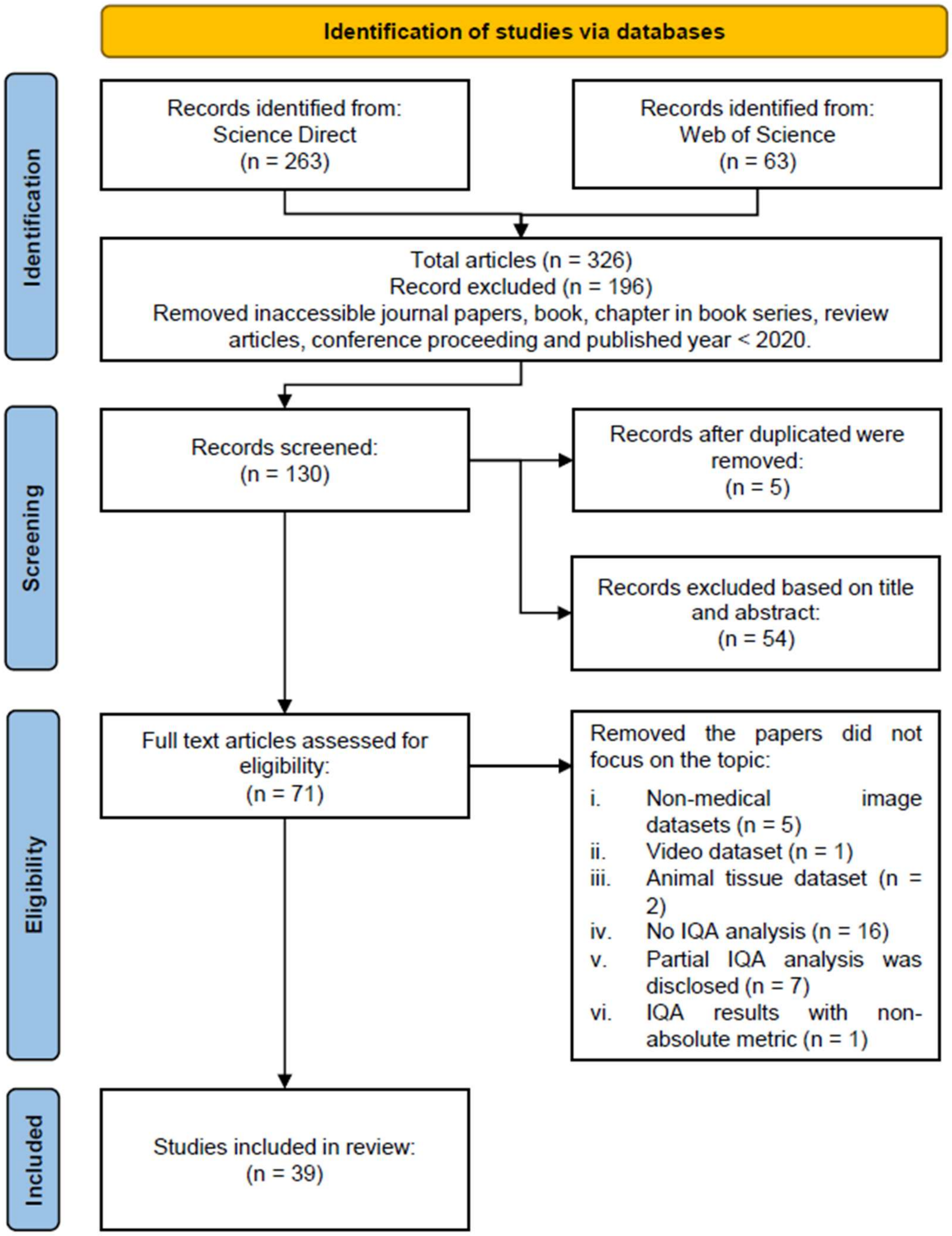

Figure 3.1 PRISMA Flow diagram for systematic review article selection

### 3.2 Quality Assessment Results of the Reviewed Articles

The quality scores of the 39 reviewed articles are summarized in Table 3.1, ranging from 75% to 95%, which highlights the overall high standard of the analyzed studies. Articles with scores of 85% or higher are categorized as good quality, as they effectively met most of the evaluation criteria, including clear objectives, robust research designs, and detailed presentation of results. Notably, 31 out of 39 articles achieved scores of 85% or above, demonstrating strong alignment with the assessment framework. Conversely, only two articles scored below 80%, indicating potential areas for improvement, such as providing more detailed methodologies or better addressing study limitations. The top-performing articles, with scores of 95%, stood out for their exceptional clarity, methodological rigor, and depth in discussing their objectives and conclusions, contributing significantly to the field of medical image enhancement. In summary, the findings indicate that the majority of the reviewed articles are of high quality, offering reliable and valuable insights that can drive advancements in image enhancement techniques.

Table 3.1 Quality performance scores of the analysed articles

| **Authors, Year** | **Quality Assessment Questions** | | | | | | | | | | **Overall Score** | **Overall (%)** |
|---|---|---|---|---|---|---|---|---|---|---|---|---|
| | **1** | **2** | **3** | **4** | **5** | **6** | **7** | **8** | **9** | **10** | | |
| Kandhway et al. 2020 [11] | 2 | 2 | 1 | 2 | 2 | 2 | 1 | 1 | 1 | 2 | 16/20 | 80.00 |
| Nasef et al. 2020 [12] | 2 | 1 | 2 | 2 | 2 | 2 | 1 | 2 | 1 | 2 | 17/20 | 85.00 |
| Subramani et al. 2020 [13] | 2 | 1 | 2 | 2 | 2 | 2 | 1 | 2 | 1 | 2 | 17/20 | 85.00 |
| Siracusano et al. 2020 [14] | 2 | 2 | 1 | 2 | 2 | 2 | 1 | 2 | 1 | 2 | 17/20 | 85.00 |
| Acharya et al. 2021 [15] | 2 | 2 | 1 | 2 | 2 | 2 | 1 | 1 | 1 | 2 | 16/20 | 80.00 |
| Rawat et al. 2021 [16] | 2 | 2 | 2 | 2 | 2 | 2 | 2 | 2 | 1 | 2 | 19/20 | 95.00 |
| Cao et al. 2021 [17] | 2 | 2 | 2 | 2 | 2 | 1 | 1 | 2 | 1 | 2 | 17/20 | 85.00 |
| Kumar et al. 2021 [18] | 2 | 2 | 1 | 2 | 2 | 2 | 2 | 2 | 1 | 2 | 18/20 | 90.00 |
| Voronin et al. 2021 [19] | 2 | 2 | 1 | 2 | 2 | 1 | 2 | 2 | 1 | 2 | 17/20 | 85.00 |
| Jalab et al. 2021 [20] | 2 | 2 | 2 | 2 | 2 | 2 | 2 | 2 | 1 | 2 | 19/20 | 95.00 |
| Kumar et al. 2022 [21] | 2 | 2 | 1 | 2 | 2 | 2 | 2 | 2 | 1 | 2 | 18/20 | 90.00 |
| Ghosh et al. 2022 [22] | 2 | 2 | 1 | 2 | 2 | 2 | 1 | 2 | 1 | 2 | 17/20 | 85.00 |
| Huang et al. 2022 [23] | 2 | 2 | 1 | 2 | 2 | 2 | 1 | 2 | 1 | 2 | 17/20 | 85.00 |
| Kumar et al. 2022 [24] | 2 | 1 | 2 | 2 | 2 | 1 | 1 | 2 | 1 | 2 | 16/20 | 80.00 |
| Kaur et al. 2022 [25] | 2 | 2 | 2 | 2 | 2 | 2 | 2 | 2 | 1 | 2 | 19/20 | 95.00 |
| Liu et al. 2022 [26] | 2 | 1 | 1 | 2 | 2 | 1 | 1 | 2 | 1 | 2 | 15/20 | 75.00 |
| Ibrahim et al. 2022 [27] | 2 | 2 | 2 | 2 | 2 | 2 | 2 | 2 | 1 | 2 | 19/20 | 95.00 |
| Sharif et al. 2022 [28] | 2 | 2 | 2 | 2 | 2 | 2 | 2 | 2 | 1 | 2 | 19/20 | 95.00 |
| Karim et al. 2022 [29] | 2 | 2 | 1 | 2 | 2 | 2 | 1 | 2 | 1 | 2 | 17/20 | 85.00 |
| Abdel-Basset et al. 2022 [30] | 2 | 2 | 1 | 2 | 2 | 1 | 2 | 1 | 1 | 2 | 16/20 | 80.00 |
| Navaneetha Krishnan et al. 2022 [31] | 2 | 2 | 1 | 2 | 2 | 2 | 2 | 1 | 1 | 2 | 17/20 | 85.00 |
| Mouzai et al. 2023 [32] | 2 | 2 | 2 | 2 | 2 | 2 | 2 | 2 | 1 | 2 | 19/20 | 95.00 |
| Wu et al. 2023 [33] | 2 | 2 | 2 | 2 | 2 | 2 | 2 | 2 | 1 | 2 | 19/20 | 95.00 |
| Ben-Loghfyry et al. 2023 [34] | 2 | 2 | 1 | 2 | 1 | 1 | 2 | 2 | 1 | 2 | 16/20 | 80.00 |
| Sule et al. 2023 [35] | 2 | 2 | 2 | 2 | 2 | 2 | 2 | 1 | 1 | 2 | 18/20 | 90.00 |
| Rao et al. 2023 [36] | 2 | 2 | 1 | 2 | 2 | 2 | 2 | 2 | 1 | 2 | 18/20 | 90.00 |
| Okuwobi et al. 2023 [37] | 2 | 2 | 2 | 2 | 2 | 2 | 2 | 2 | 1 | 2 | 19/20 | 95.00 |
| Yu et al. 2023 [38] | 2 | 2 | 2 | 2 | 2 | 2 | 2 | 2 | 1 | 2 | 19/20 | 95.00 |
| Jiang et al. 2023 [39] | 2 | 2 | 2 | 2 | 2 | 2 | 2 | 2 | 1 | 2 | 19/20 | 95.00 |
| Pashaei et al. 2023 [40] | 2 | 1 | 1 | 2 | 2 | 2 | 2 | 1 | 1 | 2 | 16/20 | 80.00 |
| Zhong et al. 2023 [41] | 2 | 2 | 1 | 2 | 2 | 1 | 2 | 2 | 1 | 2 | 17/20 | 85.00 |
| Mousania et al. 2023 [42] | 2 | 2 | 2 | 2 | 2 | 2 | 2 | 2 | 1 | 2 | 19/20 | 95.00 |
| Trung 2023 [43] | 2 | 1 | 1 | 2 | 2 | 1 | 1 | 1 | 2 | 2 | 15/20 | 75.00 |
| Jiang et al. 2024 [44] | 2 | 2 | 2 | 2 | 2 | 2 | 2 | 2 | 1 | 2 | 19/20 | 95.00 |
| Guo et al. 2024 [45] | 2 | 2 | 2 | 2 | 2 | 1 | 2 | 2 | 1 | 2 | 18/20 | 90.00 |
| Acharya et al. 2024 [46] | 2 | 2 | 1 | 2 | 2 | 2 | 1 | 2 | 1 | 2 | 17/20 | 85.00 |
| Xu et al. 2024 [47] | 2 | 2 | 1 | 2 | 2 | 2 | 1 | 2 | 1 | 2 | 17/20 | 85.00 |
| Chandra et al. 2024 [48] | 2 | 2 | 2 | 2 | 2 | 2 | 2 | 2 | 1 | 2 | 19/20 | 95.00 |
| Cap et al. 2025 [49] | 2 | 2 | 2 | 2 | 2 | 2 | 2 | 2 | 1 | 2 | 19/20 | 95.00 |

### 3.3 Challenges with Image Quality in Modern Medical Imaging Modalities

Medical imaging modalities have become indispensable tools in clinical diagnostics, offering insights into complex medical conditions. However, one significant challenge lies in ensuring optimal image quality, as poor-quality images can hinder accurate diagnosis and analysis. This section reviews 39 studies to identify and analyse the prevalent image quality issues encountered across various medical imaging modalities. The reviewed studies highlight several common image quality issues, such as low contrast, noise, brightness inconsistencies, uneven illumination, blurring, artifacts, and colour imbalance. These issues affect the interpretability of images and can significantly influence the performance of downstream analysis and diagnostic systems.

To provide a clearer understanding of the prevalence of these issues, Figure 3.2 presents a bar chart that provides a clearer representation of the frequency of each image quality issue, helping to visualize the prevalence of the mentioned problems. This chart highlights the areas that require attention to improve diagnostic accuracy in medical imaging. The dataset emphasizes the frequency of various image quality issues encountered in medical imaging across the 39 reviewed papers. Among these issues, low contrast is the most prevalent, occurring 33 times, accounting for most reported problems. This suggests that contrast-related issues are a common challenge in medical imaging, possibly hindering accurate interpretation. Noise follows with 15 occurrences, indicating its significant impact on image clarity and diagnostic performance. Brightness inconsistencies were noted 7 times, and uneven illumination was found 8 times, both affecting image consistency and potentially complicating analysis. Blurring appeared 5 times, indicating challenges in achieving sharp and detailed images. Artifacts were reported 3 times, highlighting distortions that can interfere with proper image interpretation. Finally, colour imbalance was the least frequent issue, appearing in only 2 instances.

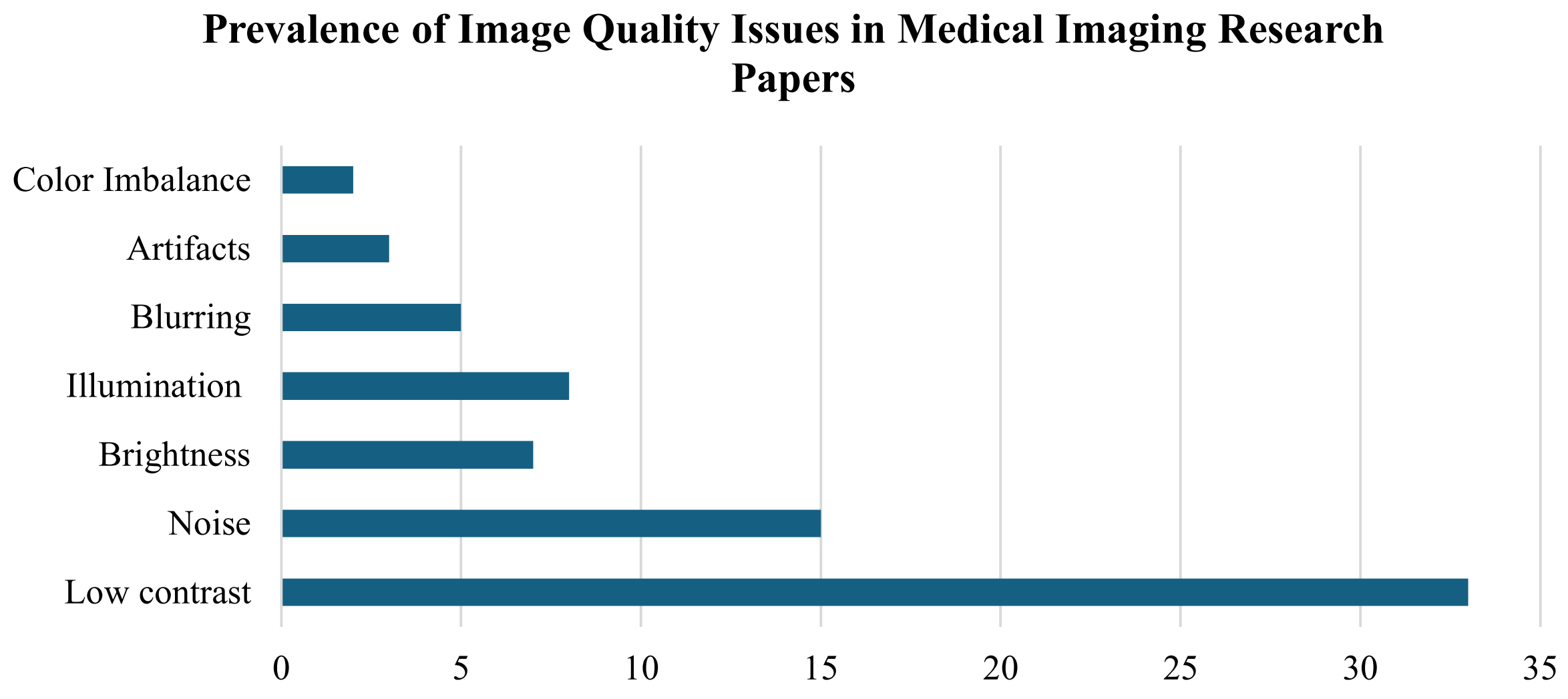

| | Low contrast | Noise | Brightness | Illumination | Blurring | Artifacts | Color Imbalance |
|---|---|---|---|---|---|---|---|
| Frequency | 33 | 15 | 7 | 8 | 5 | 3 | 2 |

Figure 3.2 Analysis of image quality deficiencies in reviewed medical imaging papers

Figure 3.3 further contextualizes these images quality challenges by illustrating the diverse distribution of imaging modalities across the 39 reviewed papers. Multi-modal imaging leads with 30.8%, followed by MRI at 17.9%, and X-ray & mammogram at 15.4%. Retinal and microscopy imaging both account for 10.3%, while CT scans contribute 7.7%. Endoscopy and bone scintigraphy imaging represent the smallest share, with 5.1% and 2.6%, respectively. This distribution underscores the prominent focus on MRI and multi-modal approaches, while less emphasis is placed on more specialized techniques like endoscopy and bone scintigraphy.

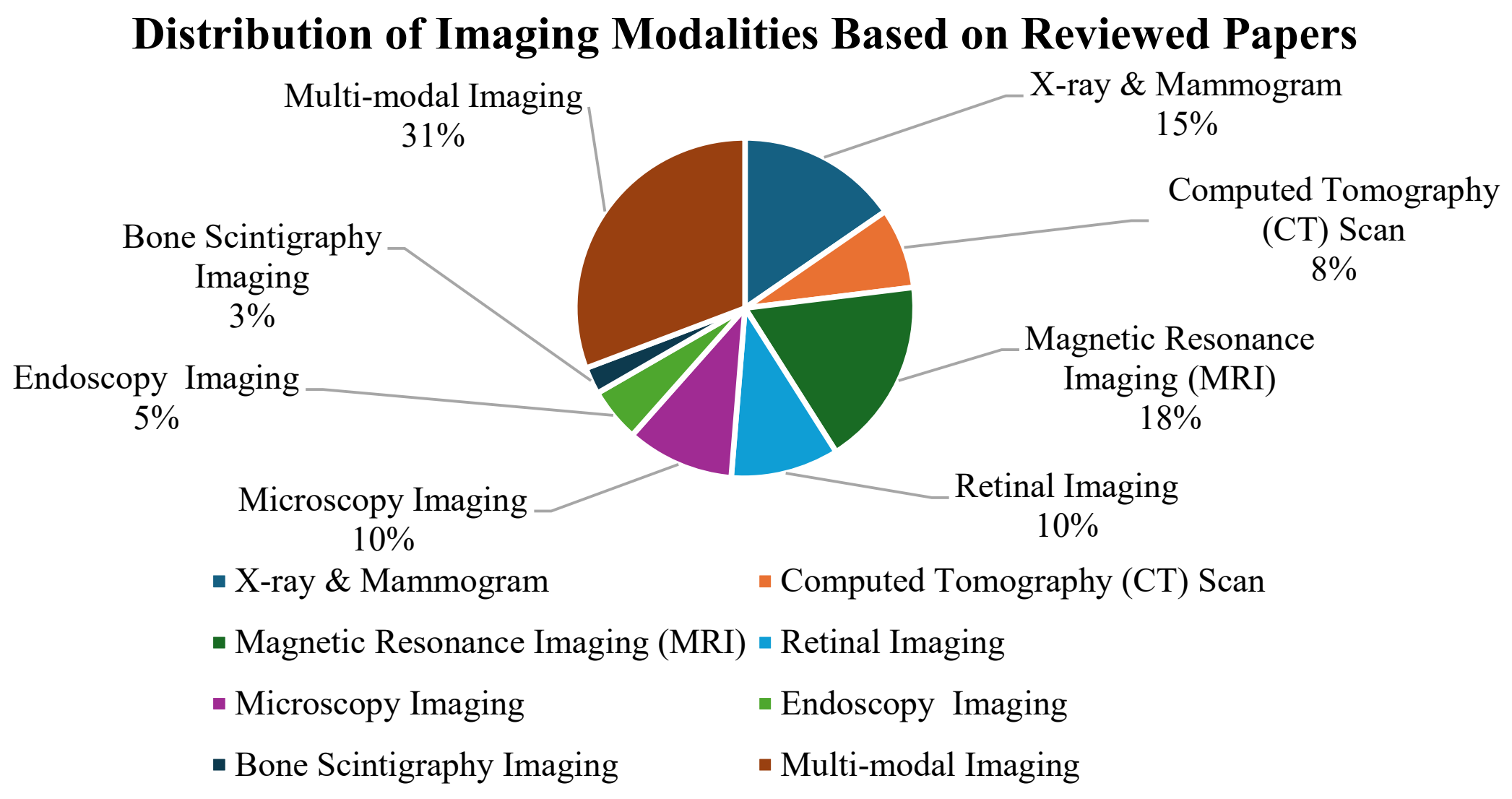

Figure 3.3 Paper distribution across different imaging modalities

To further enhance the understanding of these issues, Table 3.2 summarizes the frequency of the image quality issues identified across the 39 papers. Meanwhile, Table 3.3 presents a summary of the types of images and the associated dataset links or sources, which are invaluable for future research. These tables are closely linked and will be analysed together in the following discussion.

A. X-ray and Mammogram

Image quality issues like noise, low contrast, poor illumination, and artifacts are significant challenges in X-rays and mammograms, impacting diagnostic accuracy. Studies emphasize that these problems hinder the detection of critical features such as lesions, fractures, and pathological patterns, particularly in diverse datasets and clinical scenarios. For example, Siracusano et al. (2020) and Rawat et al. (2021) highlighted noise and low contrast in chest X-rays (CXRs), underscoring their role in obscuring diagnostic details, especially in hospital and pediatric settings [14], [16]. In the following year, Ghosh et al. (2022) focused on poor illumination and contrast in mammograms and X-rays, advocating adaptive enhancement techniques to improve visibility [22]. In the same year, Liu et al. (2022) and Abdel-Basset et al. (2022) demonstrated the utility of AI-driven noise reduction and contrast augmentation in digital radiography, particularly for COVID-19 diagnosis [26], [30]. In light of these ongoing issues, Mouzai et al. (2023) stressed the need for standardized imaging protocols to address low contrast in spine and hand X-rays [32].

B. Computed Tomography (CT) Scan

CT imaging faces challenges like low contrast, noise, and brightness issues, which impact diagnostic accuracy. Studies on CT scans reveal how these quality issues can obscure critical anatomical features and hinder accurate diagnoses. In 2022, Kaur et al. highlighted low contrast and brightness in CT scans from a private Indian dataset, suggesting the need for enhanced contrast enhancement techniques [25]. In contrast, Jiang et al. (2023) focused on noise, low contrast, and brightness in CT scans of acute appendicitis, advocating for denoising and brightness normalization to improve image clarity [39]. Similarly, Rao et al. (2023) examined CT images from the CTisus and Radpod databases, highlighting similar issues and recommending noise reduction and higher-resolution imaging for better diagnostic outcomes [36].

C. Magnetic Resonance Imaging (MRI)

The quality of medical images, especially MRI, directly impacts diagnostic accuracy. Despite advancements, issues like noise, low contrast, uneven brightness, and artifacts persist. In 2020, Subramani et al. (2020) emphasized the limitations of datasets like Radiology Assistant, MR-TIP, and BrainWeb, noting their inability to replicate real-world variability in noise and contrast [13]. Similarly, Acharya et al. (2021) and Pashaei et al. (2023) highlighted low-contrast challenges in MRI datasets such as MedPix and MRTIP, calling for adaptive enhancement techniques to address intensity variability [15], [40].

Focussing on different quality issues, Kumar et al. (2022) addressed challenges of low contrast and uneven brightness in datasets of healthy and diseased brains by introducing preprocessing techniques such as the spatial mutual information (SMI) method to enhance tumour segmentation [24]. In contrast, Ben-Loghfyry et al. (2023) focused on noise in brain MRI images, proposing an extended Perona-Malik framework for denoising while preserving anatomical details [34]. Meanwhile, Trung (2023) tackled issues of contrast and brightness inconsistencies in the AANLIB dataset by Harvard Medical School [43]. Additionally, Jiang et al. (2024) explored methods to eliminate thermal noise in datasets like BraTS 2018, further advancing imaging accuracy [44].

## D. Retinal Imaging

Retinal imaging is essential for diagnosing ocular and systemic diseases, but issues like low contrast, uneven brightness, blurring, artifacts, and colour imbalances can undermine image quality and diagnostic accuracy. One of the examples referring to study by Cao et al. (2021) addressed issues like low contrast, blurriness, and uneven brightness in retinal images, using datasets from both handheld and high-end devices, along with a private dataset from the Beijing Institute of Ophthalmology and Tongren Hospital [17]. Similarly, Kumar et al. (2022) focused on uneven illumination and contrast in retinal images from the STARE dataset [21]. Sule et al. (2023) further explored the similar issues mentioned by Kumar et al. (2022), alongside colour imbalance, using five different retinal imaging datasets [35]. In another study, Guo et al. (2024) examined synthetic and real-world retinal fundus images and identified challenges like uneven illumination, artifacts, and blurring using the EyeQ, DRIVE, and REFUGE datasets [45]. Together, these studies underscore the importance of addressing image quality challenges with strategies like illumination correction, colour normalization, and innovative machine learning techniques to enhance diagnostic accuracy and reliability in retinal imaging.

## E. Microscopy Imaging

Microscopy imaging is crucial for detailed medical diagnostics but faces challenges such as low contrast, noise, uneven brightness, and blurring, which can compromise diagnostic accuracy. Studies have explored these issues in different microscopy applications, particularly with images from the CORN-2 dataset. In 2023, Wu et al. identified uneven brightness, low contrast, and blurriness in nailfold capillary microscopy images, which hinder the visualization of fine vascular structures [33]. On the contrary, Yu et al. (2023) and Zhong et al. (2023) both focused on corneal confocal microscopy images from the CORN-2 dataset, noting problems with contrast, heterogeneous illumination, and speckle noise [38], [41]. Zhong et al. emphasized how these artifacts interfere with the visualization of corneal layers [41], while Xu et al. (2024) further highlighted the negative impact of noise on the clarity of cellular structures [47]. These studies collectively emphasize the need for addressing these challenges to improve the diagnostic value of microscopy images, particularly in datasets like CORN-2, to enhance the visualization of cellular and vascular features.

### F. Endoscopy and Bone Scintigraphy Imaging

Endoscopy and skeletal scintigraphy are vital imaging modalities in medical diagnostics but face significant challenges that impact diagnostic precision. Studies have identified key issues such as low contrast, noise, illumination inconsistencies, blurring, and color deviations. For instance, Huang et al. (2022) examined gastrointestinal endoscopic images from datasets including Kvasir, Kvasir-SEG, CVC-ClinicDB, and ETIS-Larib Polyp DB, highlighting problems such as low illumination, poor brightness, and color deviations that complicate the identification of polyps and other abnormalities [23]. Similarly, Cap et al. (2025) focused on endoscopic throat images from a private dataset, identifying blurring, low contrast, and uneven illumination, which hindered the assessment of throat conditions like tumours or inflammation [49]. In skeletal scintigraphy, Nasef et al. (2020) studied low-contrast issues in images from a private dataset at Menoufia University Hospital [12]. These studies collectively highlight persistent issues in both endoscopy and skeletal scintigraphy, emphasizing the need for standardized imaging protocols and improvements in image quality to enhance diagnostic accuracy.

### G. Multi-modal Imaging

Multi-modal imaging, which involves various imaging modalities like CT, MRI, X-ray, and ultrasound, plays a crucial role in modern medical diagnostics but faces persistent challenges in image quality, including low contrast, noise, blurring, and artifacts, which hinder diagnostic accuracy. In 2020, Kandhway et al. analyzed mammograms, X-rays, MRIs, and CT scans from the MIAS and LITFL datasets, highlighting low contrast as a key issue, especially in dense tissues like mammography [11]. Similarly, Jalab et al. (2021) explored lung CT, brain MRI, and kidney MRI images from COVID-19 and brain datasets, noting that poor contrast made it difficult to identify pathologies [20]. In the same year, Voronin et al. (2021) addressed blur and low contrast, proposing adaptive deblurring to mitigate motion artifacts in fastMRI datasets [19].

Following this, Kumar et al. (2021), Ibrahim et al. (2022), and Karim et al. (2022) explored low contrast issues in CT and X-ray images from COVID-19-related datasets [18], [27], [29]. Kumar et al. highlighted how low contrast in the COVID-19 CT and X-ray dataset made it difficult to distinguish COVID-19-related changes [18]. Similarly, Ibrahim et al. found that low contrast in both CT and MRI images from the COVID-19 CT and Brain MRI datasets obscured critical features in lung and brain scans [27]. Karim et al. also emphasized similar challenges with low contrast in chest X-ray and CT scans from the COVID-19 Chest X-ray and Italian Society databases, complicating the detection of subtle abnormalities [29]. Together, these studies underscore the persistent problem of low contrast in COVID-19 imaging datasets, affecting diagnostic accuracy across modalities.

Furthermore, artifact-related issues were highlighted by Sharif et al. (2022), who examined MRI, X-ray, skin, and protein atlas images across multiple databases [28]. They noted that artifacts, such as motion and metal artifacts, introduced noise that complicated image interpretation. In a similar vein, Navaneetha Krishnan et al. (2022) pointed out that noise and low contrast impacted CT, MRI, and dermoscopic images, affecting the clarity of diagnostic features [31]. Continuing the theme, Okubowi et al. (2023) and Chandra et al. (2024) emphasized noise and low contrast issues in various imaging modalities [37], [48]. These include X-ray, CT, and retinal vascular images, as well as MRI and ultrasound, which impact applications such as tumor detection and neurological assessments. Lastly, Mousania et al. (2023) and Acharya et al. (2024) examined low contrast and artifacts across various imaging modalities, including mammograms, ultrasound, MRI, and CT scans [42], [46]. These studies collectively underscore the ongoing challenges of low contrast, noise, blurring, and artifacts in multi-modal imaging, stressing the need for continuous advancements in preprocessing, standardization, and optimization to improve diagnostic accuracy and consistency across diverse imaging modalities.

### 3.4 Advancement of Recent Medical Image Enhancement Approaches

Image enhancement techniques can be broadly classified into conventional methods and deep learning-based approaches. Out of 39 studies, 29 continue to incorporate conventional concepts, integrating advanced mathematical techniques to refine algorithm development. Meanwhile, 9 studies focus on deep learning approaches, and 1 study explores a hybrid method that combines both conventional and deep learning techniques. To provide a comprehensive analysis of the methodologies, all the methods proposed by recent studies have been summarized in Table 3.4, along with their results, advantages, limitations, and the software used.

Among the conventional techniques, many studies focused on contrast enhancement using histogram-based methods. Several studies [13], [15], [18], [42] applied various forms of histogram equalization, with modifications such as fuzzy logic-based adaptive histogram equalization [13], genetic algorithm-optimized histogram equalization [15], weighted histogram equalization with gamma correction [24], and a hybrid approach merging direct and indirect histogram equalization techniques [42]. Additionally, [11], [12], [20], [22], [27], [29], [30] leveraged fractional calculus, entropy concepts, or geometric functions to enhance contrast, demonstrating an alternative mathematical perspective in improving image quality.

Apart from contrast enhancement, post-processing techniques were another area of study, with [14], [26], [34] introducing wavelet-based and multiscale noise reduction techniques to suppress artifacts while enhancing important image details. Notably, [14] integrated Fast and Adaptive Bidimensional Empirical Mode Decomposition (FABEMD), Homomorphic Filtering (HMF), and Contrast Limited Adaptive Histogram Equalization (CLAHE) in a post-processing pipeline to improve chest X-ray quality. Similarly, [26] utilized Shannon-Cosine wavelets for multiscale noise reduction, while [34] incorporated time-fractional derivatives and adaptive diffusion to restore images effectively. Other studies [31], [33], [35], [36] addressed noise reduction by modifying median filtering [31], applying non-local means filtering [33], optimizing CLAHE parameters [35], and integrating wavelet-based techniques with adaptive morphology [36].

Moreover, bio-inspired and metaheuristic algorithms were also widely explored for optimization in image enhancement. Studies [11], [12], [40] applied nature-inspired techniques such as krill herd optimization, bio-inspired swarm algorithms, and metaheuristic approaches to optimize enhancement parameters dynamically. Similarly, [39], [40], [46] introduced metaheuristic algorithms to improve contrast and denoising performance, making optimization a key aspect of enhancement strategies. Additionally, [37] proposed a heuristic optimization approach based on a novel local transfer function to enhance image quality.

In recent years, deep learning (DL)-based techniques have gained significant traction in recent years. Studies [16], [28], [32], [38], [41], [44], [45], [47], [49] explored various DL-based frameworks for image enhancement. Several studies [16], [28], [32] proposed CNN-based approaches, such as residual learning [16] and attention mechanisms [32]. Generative adversarial networks (GANs) were another prominent DL method, with [38], [41], [45], [47], [49] integrating GANs for image enhancement. Specifically, combined fuzzy theory with adversarial learning to correct illumination, while [41] adopted an attention-based GAN enhancement method. In addition, [44], [45], [47] focused on network improvements, including ARM-Net for thermal noise removal [44], a multi-degradation-adaptation module using GAN [45], and a dual-input Siamese network for structure-preserving enhancement [47]. Moreover, [49] introduced an unsupervised GAN-based method leveraging Laplacian theory to handle blurry images. In a different approach, [23] proposed a deep unsupervised learning framework based on a multi-image fusion method along with conventional methods.

Apart from individual techniques, some studies took a fusion-based approach, integrating multiple enhancement techniques for superior results. Studies [17], [18], [19] built upon motivations in [13], refining contrast enhancement through fusion-based techniques, such as optimizing channel selection [17], achieving brightness preservation [18], and implementing a 3D block-rooting scheme optimized using the Golden transform [19]. Similarly, [25], [36] explored fusion-based filtering techniques, where [25] applied an anisotropic diffusion filter combined with windowing techniques, and [36] incorporated wavelet-based and adaptive morphology for enhancement. Beyond these established categories, some research works proposed novel enhancement mechanisms that do not fit within traditional categories. Study [43] applied a fuzzy logic-based clustering method for contrast enhancement, while [48] relied on Type II fuzzy membership functions and the Hamacher T-conorm operator.

In analyzing these 39 studies, MATLAB is the most commonly used software, with 14 studies utilizing various versions (e.g., MATLAB 2018/2019, MATLAB 2017a, or general versions) [12], [15], [20], [21], [23], [25], [29], [30], [31], [34], [37], [40], [46], [48]. Python-based tools, such as OpenCV, TensorFlow, Keras, and PyTorch, are employed in 5 studies. Additionally, hardware setups are specified in 10 studies, most notably involving NVIDIA GPUs. Thirteen studies do not mention the software used, which stands out as a significant number compared to the studies that specify tools. Other mentioned tools include OpenCV, ArrayFire, Scikit-image, Google Colab, NumPy, and CentOS Linux, each appearing in a single study.

Overall, the literature demonstrates a clear shift from conventional enhancement techniques toward AI-driven and hybrid approaches. Optimization, noise suppression, and contrast enhancement remain key research themes, with deep learning methods increasingly dominating the field. These advancements provide a strong foundation for future work in medical image enhancement, particularly in applications requiring high-precision imaging.

### 3.5 Image Quality Assessment (IQA)

This section provides a comprehensive analysis of the image quality assessment (IQA) metrics proposed in the reviewed studies. These metrics are categorized into reference-based and non-reference-based IQA methods, as outlined in Tables 3.5 and 3.6, respectively. Each table presents the concept and mathematical formulation of the proposed metrics, along with their corresponding indications of image quality. Specifically, a value of ‘1’ signifies that a higher metric value reflects better image quality, whereas a value of ‘0’ indicates that a lower metric value corresponds to higher image quality.

Among the 39 studies reviewed, 18 employed both reference-based and non-reference-based metrics [11], [13], [14], [15], [17], [18], [21], [23], [28], [31], [32], [33], [35], [36], [37], [40], [42], [46], while 9 studies relied solely on reference-based metrics [16], [22], [25], [26], [30], [34], [44], [45], [48], and 12 exclusively utilized non-reference-based metrics [12], [19], [20], [24], [27], [29], [38], [39], [41], [43], [47], [49] to evaluate their proposed algorithms. In total, 65 distinct IQA metrics were introduced across these studies, with a significant majority being non-reference-based. Specifically, 42 of the metrics were non-reference-based, while 23 were reference-based. Notably, 13 metrics were associated with a ‘0’ indication, whereas 52 metrics were denoted with ‘1’, suggesting that most IQA methods favor higher values to indicate superior image quality.

An emerging trend observed in these studies is the increasing integration of deep learning-based IQA metrics, which offer enhanced perceptual quality assessment capabilities. Among the 6 deep learning-based metrics identified, one reference-based metric, Learned Perceptual Image Patch Similarity (LPIPS) [50], has gained popularity for its ability to capture perceptual differences effectively. Meanwhile, five non-reference-based deep learning metrics have been introduced: Neural Image Assessment (NIMA) [51], From Patches to Pictures (PaQ-2-PiQ) [52], Deep bilinear convolutional neural network (DBCNN) [53], HyperIQA [54] and Multi-scale Image Quality (MUSIQ) [55]. These methods leverage deep neural networks to assess image quality in a more human-like manner, making them particularly useful for real-world applications where ground truth references are unavailable. The prevalence of non-reference-based deep learning metrics highlights a shift towards more automated and adaptive IQA techniques, capable of evaluating complex distortions beyond traditional handcrafted methods.

In line with this trend, two novel no-reference image quality metrics have been introduced: the Golden Image Quality Enhancement Measure (GIQEM) and the Laplacian Structural Similarity Index Measure (LaSSIM), proposed in studies [19] and [49], respectively. GIQEM measures contrast enhancement using the Golden transform by capturing high-frequency content, making it particularly useful for evaluating enhancement techniques. Meanwhile, LaSSIM assesses the structural preservation of medical images by applying Laplacian Pyramid (LP) decomposition before computing the Structural Similarity Index Measure (SSIM), ensuring a more refined evaluation of structural integrity. These novel metrics further reinforce the growing emphasis on non-reference-based IQA approaches, particularly in medical imaging, where reference images may not always be available.

## 4.0 Discussion

Medical imaging quality plays a pivotal role in clinical diagnostics, directly influencing the interpretability and accuracy of diagnostic systems. A review of 39 studies reveals persistent challenges such as low contrast, noise, blurring, uneven brightness, artifacts, and color imbalance. Among these, low contrast is the most frequently reported issue across various imaging modalities, followed closely by noise, which further complicates image clarity and interpretability.

The impact of these challenges varies by modality. In X-rays and mammograms, noise, low contrast, and brightness inconsistencies obscure critical diagnostic features such as lesions and fractures. Similarly, CT scans suffer from brightness inconsistencies and noise, making anatomical visualization difficult. Despite continuous technological advancements, MRI remains prone to artifacts, uneven brightness, and low contrast, often due to the limitations of datasets in replicating real-world variability. Retinal and microscopy imaging, essential for ocular and cellular-level diagnostics, experience uneven illumination, blurring, and artifacts, which hinder accurate analysis. Furthermore, specialized imaging techniques such as endoscopy and bone scintigraphy face low contrast and blurring, reducing diagnostic precision. A notable trend in research is the strong focus on MRI and multi-modal imaging (48.7%), whereas specialized modalities such as bone scintigraphy and endoscopy remain underexplored. Additionally, histopathological imaging, crucial for cancer diagnosis, is insufficiently addressed, despite its unique challenges, including staining-induced color variability, uneven illumination, and high sensitivity to noise.

The review of recent medical image enhancement techniques highlights a transition from traditional mathematical approaches to deep learning-based methods, with hybrid models gaining traction. Conventional techniques, particularly histogram equalization and noise reduction methods, remain widely used due to their interpretability and mathematical rigor. However, deep learning approaches, including convolutional neural networks (CNNs) and generative adversarial networks (GANs), have demonstrated superior performance in handling complex imaging conditions. Furthermore, the integration of optimization algorithms, such as metaheuristic techniques, has enhanced enhancement strategies by dynamically adjusting parameters. Notably, fusion-based methods, which combine multiple enhancement techniques, have shown promising results in balancing contrast improvement, noise suppression, and structure preservation. Despite this methodological diversity, a lack of standardization in software usage and benchmarking across studies remains a critical limitation. While MATLAB is the predominant tool in conventional studies, deep learning-based approaches rely on Python frameworks such as TensorFlow and PyTorch. However, many studies omit software details altogether, hindering reproducibility and comparative analysis. The shift toward AI-driven enhancement underscores its potential to improve medical imaging quality, ultimately enabling more precise diagnostics and clinical decision-making.

Similarly, the review of image quality assessment (IQA) metrics highlights a growing shift from traditional reference-based methods to more adaptive non-reference-based approaches. This transition is particularly relevant in medical imaging, where ground truth references are often unavailable. Among the 65 identified IQA metrics, 42 are non-reference-based, reflecting the increasing need for independent evaluation techniques. Deep learning-based IQA methods have gained significant traction, demonstrating superior perceptual quality assessment capabilities compared to handcrafted metrics. The adoption of learned perceptual models, such as LPIPS, NIMA, and HyperIQA, further signifies the field's reliance on AI-driven evaluation techniques. Additionally, the introduction of novel domain-specific IQA measures, such as the Golden Image Quality Enhancement Measure (GIQEM) and the Laplacian Structural Similarity Index Measure (LaSSIM), highlights the need for specialized assessment tools tailored to medical image enhancement. However, the wide variation in IQA metrics across studies points to a lack of standardization, posing challenges for consistent benchmarking and cross-study comparisons. Overall, the increasing adoption of deep learning-based and non-reference-based IQA methods represents a crucial transformation in medical image assessment, promoting more accurate and perceptually meaningful evaluations.

**5.0 Research Gaps and Future Directions**

Despite progress in medical image enhancement, several research gaps remain. Specialized imaging modalities like bone scintigraphy, endoscopy, and histopathology require more attention, particularly in addressing staining variability, colour imbalance, and noise. AI-driven methods show promise but lack seamless integration into standardized imaging pipelines. Additionally, existing datasets often fail to capture real-world clinical variability, limiting the effectiveness of AI solutions.

Future research should focus on adaptive algorithms for contrast enhancement and noise reduction, particularly in underexplored modalities. Generative adversarial networks (GANs) could improve staining normalization in histology. Open-access datasets reflecting real-world variability and standardized image quality benchmarks would enhance reliability. Cross-modality preprocessing solutions should be developed to unify AI-driven enhancements across different imaging domains. Additionally, explainable AI (XAI) can increase transparency in automated image processing, especially in cancer detection. Standardized imaging protocols across institutions are essential for improving diagnostic consistency.

Deep learning models often rely on large, labelled datasets, which are scarce in medical imaging. Developing self-supervised or unsupervised learning models can mitigate this limitation. While many enhancement methods improve contrast, they may introduce artifacts or degrade essential diagnostic details. Hybrid approaches should balance enhancement and structural preservation. Standardized evaluation metrics and benchmark datasets would improve performance comparisons. Real-time deployment remains challenging, particularly in clinical settings where computational efficiency is critical. Lightweight AI models optimized for real-time edge-device processing should be prioritized.

Image quality assessment (IQA) also faces unresolved challenges. Non-reference-based metrics, while practical, often lack well-defined ground truth validation. Future IQA models should integrate statistical and deep learning-based perceptual assessments. Current deep learning-based IQA methods are mostly derived from natural image datasets and do not fully capture medical image distortions. Large-scale medical IQA datasets are needed for better training. Standardization is another key issue, as varying metrics hinder cross-study comparisons. Establishing benchmark datasets and evaluation protocols would enhance reproducibility. Computational efficiency should be prioritized for real-time clinical applications, requiring lightweight and interpretable IQA frameworks.

In summary, AI-driven medical image enhancement and IQA have advanced significantly, but challenges remain. Future work should focus on adaptive algorithms, standardized evaluation, real-world datasets, and real-time implementation to improve clinical applicability.

## 5.0 Conclusion

This systematic literature review highlights the significant progress made in medical image enhancement and quality assessment, particularly with the adoption of AI-driven methods such as deep learning. Despite notable advancements, several challenges remain, particularly in specialized imaging modalities like bone scintigraphy, endoscopy, and histopathology, where issues like staining variability and noise are prevalent. Additionally, the lack of standardized evaluation metrics and the scarcity of real-world clinical datasets hinder the development of universally applicable solutions. Future research should focus on adaptive algorithms for contrast enhancement, noise reduction, and the integration of AI models into standardized imaging pipelines. Moreover, the creation of open-access datasets and the establishment of standardized IQA metrics and evaluation protocols will enhance the reproducibility and applicability of medical image enhancement techniques. By tackling these challenges, the field can better support accurate and efficient healthcare solutions, ultimately contributing to improved patient outcomes.

Table 3.2 Frequency of image quality issues identified in recent studies

| Authors, Year | Types of Images | Low contrast | Noise | Brightness | Illumination | Blurring | Artifacts | Color Imbalance |
|---|---|---|---|---|---|---|---|---|
| Kandhway et al. 2020 [11] | Mammogram, X-ray, MRI, and CT scan images from different body parts | ✓ | | | | | | |
| Nasef et al. 2020 [12] | Skeletal scintigraphy images | ✓ | | | | | | |
| Subramani et al. 2020 [13] | MRI images | ✓ | ✓ | | | | | |
| Siracusano et al. 2020 [14] | Chest X-rays (CXRs) | ✓ | ✓ | | | | | |
| Acharya et al. 2021 [15] | MRI scans | ✓ | | | | | | |
| Rawat et al. 2021 [16] | X-ray (CXR) | | ✓ | | | | | |
| Cao et al. 2021 [17] | Retinal images | ✓ | | ✓ | | ✓ | | |
| Kumar et al. 2021 [18] | CT and X-ray images | ✓ | | | | | | |
| Voronin et al. 2021 [19] | X-ray and MRI images | ✓ | | | | ✓ | | |
| Jalab et al. 2021 [20] | Lung CT and MRI images | ✓ | | | | | | |
| Kumar et al. 2022 [21] | Retinal images | ✓ | | | ✓ | | | |
| Ghosh et al. 2022 [22] | Mammogram, X-ray | ✓ | | | ✓ | | | |
| Huang et al. 2022 [23] | Endoscopic gastrointestinal tract | | | ✓ | ✓ | | | ✓ |
| Kumar et al. 2022 [24] | MRI | ✓ | | ✓ | | | | |
| Kaur et al. 2022 [25] | CT scan | ✓ | | ✓ | | | | |
| Liu et al. 2022 [26] | X-ray | ✓ | ✓ | | | | | |
| Ibrahim et al. 2022 [27] | CT and MRI | ✓ | | | | | | |
| Sharif et al. 2022 [28] | MRI, X-ray, skin and protein atlas | | | | | | ✓ | |
| Karim et al. 2022 [29] | Chest X-ray and CT scans | ✓ | | | | | | |
| Abdel-Basset et al. 2022 [30] | Chest X-ray | ✓ | ✓ | | | | | |
| Navaneetha Krishnan et al. 2022 [31] | CT, MRI and dermascopic | ✓ | ✓ | | | | | |
| Mouzai et al. 2023 [32] | X-rays | ✓ | | | | | | |
| Wu et al. 2023 [33] | Microscopy images | ✓ | | ✓ | | ✓ | | |
| Ben-Loghfyry et al. 2023 [34] | MRI images | | ✓ | | | | | |
| Sule et al. 2023 [35] | Retinal fundus image | ✓ | | | ✓ | | | ✓ |
| Rao et al. 2023 [36] | CT images | ✓ | ✓ | | | | | |
| Okuwobi et al. 2023 [37] | X-ray, CT, retinal vascular and fluorescein angiogram | ✓ | ✓ | | | | | |
| Yu et al. 2023 [38] | Corneal Confocal Microscopy images | ✓ | ✓ | | ✓ | | | |
| Jiang et al. 2023 [39] | Axial CT scans of acute appendicitis | ✓ | ✓ | ✓ | | | | |
| Pashaei et al. 2023 [40] | MRI | ✓ | | | | | | |
| Zhong et al. 2023 [41] | Corneal Confocal Microscopy images | ✓ | ✓ | | ✓ | | | |
| Mousania et al. 2023 [42] | Mammograms, ultrasound, MRI, CT scans | ✓ | | | | | | |
| Trung 2023 [43] | MRI | ✓ | | ✓ | | | | |
| Jiang et al. 2024 [44] | MRI | | ✓ | | | | | |
| Guo et al. 2024 [45] | Fundus images – Synthetic images | | | | ✓ | ✓ | ✓ | |
| Acharya et al. 2024 [46] | MRI and CT | ✓ | | | | | ✓ | |
| Xu et al. 2024 [47] | Corneal Confocal Microscopy images | ✓ | ✓ | | | | | |
| Chandra et al. 2024 [48] | MRI brain scans, X-rays and Ultrasound | ✓ | ✓ | | | | | |
| Cap et al. 2025 [49] | Endoscopic throat image | ✓ | | | ✓ | ✓ | | |
| **Frequency of Image Quality Issues** | | **33** | **15** | **7** | **8** | **5** | **3** | **2** |

Table 3.3 Summary of image types and dataset sources

| Authors, Year | Types of images | Datasets | Link / Source |
|---|---|---|---|
| Kandhway et al. 2020 [11] | Mammogram, X-ray, MRI, and CT scan images from different body parts | MIAS | https://www.mammoimage.org/databases/ |
| | | LITFL | NA |
| Nasef et al. 2020 [12] | Skeletal scintigraphy images | Private dataset | Menoufia University Hospital, Egypt |
| Subramani et al. 2020 [13] | MRI images | Radiology Assistant | https://radiologyassistant.nl/ |
| | | MR-TIP | https://www.mr-tip.com/serv1.php |
| | | BrainWeb | https://brainweb.bic.mni.mcgill.ca/brainweb/ |
| Siracusano et al. 2020 [14] | Chest X-rays (CXRs) | Private dataset | University Hospital 'Policlinico G. Martino |
| | | Public Dataset | https://github.com/ieee8023/covid-chestxray-dataset |
| Acharya et al. 2021 [15] | MRI scans | MedPix | https://medpix.nlm.nih.gov/home |
| | | OpenI | https://openi.nlm.nih.gov/faq?download=true |
| | | MRTIP | https://www.mr-tip.com/serv1.php |
| Rawat et al. 2021 [16] | X-ray (CXR) | Guangzhou Dataset from Guangzhou Women and Children's Medical Center [56] | https://data.mendeley.com/datasets/rscbjbr9sj/3 |
| Cao et al. 2021 [17] | Retinal images | Handheld Device and High-End Device | https://riadd.grand-challenge.org/Data/ |
| | | Private Dataset | Beijing Institute of Ophthalmology, Tongren Hospital |
| Kumar et al. 2021 [18] | CT and X-ray images | COVID-19 CT and X-ray image [57] | https://github.com/ieee8023/covid-chestxray-dataset |
| Voronin et al. 2021 [19] | X-ray and MRI images | fastMRI [58] | https://fastmri.med.nyu.edu/ |
| | | ChestX-ray [59] | https://nihcc.app.box.com/v/ChestXray-NIHCC |
| | | NYU [60] | https://github.com/VLOGroup/mri-variationalnetwork |
| Jalab et al. 2021 [20] | Lung CT and MRI images | COVID-19 DATABASE [61] | *https://www.sirm.org/category/senza-categoria/covid-19/* |
| | | Brain MRI [62] | Al-Kadhimiya Medical City, Iraq |
| | | Kidney MRI [63] | Hospital in Saudi Arabia |
| Kumar et al. 2022 [21] | Retinal images | STARE | http://cecas.clemson.edu/~ahoover/stare/ |
| Ghosh et al. 2022 [22] | Mammogram, X-ray | MIAS | https://www.mammoimage.org/databases/ |
| | | MedPix | https://medpix.nlm.nih.gov/home |
| | | INbreast [64] | http://medicalresearch.inescporto.pt/breastresearch/GetINbreastDatabase.html |
| | | DDSM | https://www.cancerimagingarchive.net/collection/cbis-ddsm/ |
| Huang et al. 2022 [23] | Endoscopic gastrointestinal tract | Kvasir dataset [65] | https://datasets.simula.no/kvasir/ |
| | | Kvasir-SEG [66] | https://datasets.simula.no/kvasir-seg/ |
| | | CVC-ClinicDB [67] | https://polyp.grand-challenge.org/CVCClinicDB/ |
| | | ETIS-Larib Polyp DB [68] | http://vi.cvc.uab.es/colon-qa/cvccolondb/ |
| | | CVC-EndoSceneStill [69] | https://pages.cvc.uab.es/CVC-Colon/index.php/databases/cvc-endoscenestill/ |
| | | CVC-ClinicSpec [70] | https://pages.cvc.uab.es/CVC-Colon/index.php/cvc-clinicspec/ |
| Kumar et al. 2022 [24] | MRI | Healthy brain, unhealthy brain and multiclass brain tumour | https://www.kaggle.com/datasets/sartajbhuvaji/brain-tumor-classification-mri |
| Kaur et al. 2022 [25] | CT scan | Private dataset | PGIMER, Chandigarh, India |
| Liu et al. 2022 [26] | X-ray | Digital Radiography (DR) images | NA |
| Ibrahim et al. 2022 [27] | CT and MRI | COVID-19 CT DATABASE [61] | *https://www.sirm.org/category/senza-categoria/covid-19/* |
| | | Brain MRI [71] | http://www.braintumorsegmentation.org/ |
| Sharif et al. 2022 [28] | MRI, X-ray, skin and protein atlas | Radiology – MRI [72] | https://wiki.cancerimagingarchive.net/display/Public/TCGA-LGG |
| | | Radiology – X-ray [73] | https://stanfordmlgroup.github.io/competitions/chexpert/ |
| | | Dermatology [74] | https://isic-archive.com/ ; https://www.kaggle.com/datasets/spacesurfer/ph2-dataset |
| | | Microscopy [75] | NA |
| Karim et al. 2022 [29] | Chest X-ray and CT scans | COVID-19 Chest X-ray | https://www.kaggle.com/datasets/pranavraikokte/covid19-image-dataset |
| | | COVID-19 CT DATABASE [61] | *https://www.sirm.org/category/senza-categoria/covid-19/* |
| Abdel-Basset et al. 2022 [30] | Chest X-ray | COVID-19 CXR: Normal, COVID-19, | https://www.kaggle.com/datasets/andrewmvd/convid19-x-rays |
| | | viral pneumonia and lung opacity | |
| Navaneetha Krishnan et al. 2022 [31] | CT, MRI and dermascopic | CT, MRI and Dermascopic | NA |
| Mouzai et al. 2023 [32] | X-rays | Cervical spine, lumbar spine and [76] | The second National Health and Nutrition Survey (NHANES II) - National Institutes of Health (NIH) |
| | | Hand X-rays [77] | Children's Hospital Los Angeles |
| Wu et al. 2023 [33] | Microscopy images | Nailfold capillary images | NA |
| Ben-Loghfyry et al. 2023 [34] | MRI images | MRI images from brain, skull and head | https://www.kaggle.com/datasets/ |
| Sule et al. 2023 [35] | Retinal fundus image | DRIVE [78] | https://drive.grand-challenge.org/DRIVE/ |
| | | STARE [79] | http://cecas.clemson.edu/~ahoover/stare/ |
| | | DIARETDB1 [80] | https://www.kaggle.com/datasets/nguyenhung1903/diaretdb1-v21/data |
| | | HRF [81] | https://www5.cs.fau.de/research/data/fundus-images/ |
| Rao et al. 2023 [36] | CT images | CTisus | http://www.ctisus.com/ |
| | | Radpod | *http://www.radpod.org/* |
| Okuwobi et al. 2023 [37] | X-ray, CT, retinal vascular and fluorescein angiogram | X-ray | Private Dataset |
| | | CT | |
| | | Optical Coherence Tomography Angiography (OCTA) | |
| | | Fluorescein Angiography (FA) | |
| Yu et al. 2023 [38] | Corneal Confocal Microscopy images | CORN-2 [82] | https://imed.nimte.ac.cn/CORN.html |
| Jiang et al. 2023 [39] | Axial CT scans of acute appendicitis | MedPix | https://medpix.nlm.nih.gov/home |
| Pashaei et al. 2023 [40] | MRI | MedPix | https://medpix.nlm.nih.gov/home |
| Zhong et al. 2023 [41] | Corneal Confocal Microscopy images | CORN-2 [82] | https://imed.nimte.ac.cn/CORN.html |
| Mousania et al. 2023 [42] | Mammograms, ultrasound, MRI, CT scans | MIAS | https://www.mammoimage.org/databases/ |
| | | Ultrasound Cases Database (focal liver lesions, carotid artery) [83] | https://www.ultrasoundcases.info/cases/abdomen-and-retroperitoneum/; *http://splab.cz/en/download/databaze/ultrasound* |
| | | Brain MRI | NA |
| | | CT scan | NA |
| Trung 2023 [43] | MRI | Harvard Medical School's AANLIB database | https://www.med.harvard.edu/AANLIB/ |
| Jiang et al. 2024 [44] | MRI | BraTS 2018 Dataset [71], [84], [85] | https://www.med.upenn.edu/sbia/brats2018/data.html |
| Guo et al. 2024 [45] | Fundus images – Synthetic images | EyeQ [86] | https://github.com/hzfu/EyeQ?tab=readme-ov-file |
| | | DRIVE [78] | https://drive.grand-challenge.org/DRIVE/ |
| | | REFUGE [87] | https://refuge.grand-challenge.org/ |
| Acharya et al. 2024 [46] | MRI and CT | Medpix | https://medpix.nlm.nih.gov/home ; https://openi.nlm.nih.gov/ |
| | | MRTIP | https://www.mr-tip.com/serv1.php |
| Xu et al. 2024 [47] | Corneal Confocal Microscopy images | CORN-2 [82] | https://imed.nimte.ac.cn/CORN.html |
| Chandra et al. 2024 [48] | MRI brain scans, | MRI brain scans | https://www.kaggle.com/datasets/navoneel/brain-mri-images-for-brain-tumor-detection?resource=download |
| | X-rays | X-rays [88] | https://github.com/ieee8023/covid-chestxray-dataset |
| | Ultrasound | Ultrasound [89] | NA |
| Cap et al. 2025 [49] | Endoscopic throat image | Private dataset | NA |

** Italicized text indicates that the link is either inaccessible or the file is no longer available.

** NA indicates not available.

Table 3.4 Overview of methodologies in analyzed studies

| Authors, Year | Method | Average Results | Merit | Demerit | Software / Tools |
|---|---|---|---|---|---|
| Kandhway et al. 2020 [11] | Krill herd-based and SSA-based algorithms | *SSIM = 0.8609, Edge preserve index, EPI == 1.8683, Entropy = 5.4697, relative enhancement contrast, REC = 1.0583 and fitness function = 5.0707* | Adaptive and automatic parameter optimization eliminates manual tuning and preserves critical diagnostic features like edges and texture. | High computational time. | NA |
| Nasef et al. 2020 [12] | Neutrosophic Sets (NS) and Salp Swarm Algorithm (SSA) | 512*512:<br>Fitness function = 12.00247, Entropy = 5.163514, Number of edges = 256373.1, sharpness = 99.09021, S-Index = 101.76, CEIQ =2.324094 and NIQE = 4.324512<br><br>256*256:<br>Fitness function = 11.9159, Entropy = 5.254348, number of edges = 64909.36, sharpness = 52.32788, S-Index = 53.72089, CEIQ =2.346036 and NIQE = 6.671592<br><br>128*128:<br>Fitness function = 11.97531, Entropy = 5.469861, number of edges = 16384, sharpness = 20.55806, S-Index = 21.04081, CEIQ =2.38071 and NIQE = 18.87192 | Focuses specifically on enhancing critical diagnostic regions | Performance varies with image resolution; low-resolution images (brightness 20%-35%) may result in insufficient enhancement of dark areas. | Matlab 2018a |
| Subramani et al. 2020 [13] | FGLDHE | Entropy = 7.01, PSNR = 38.15dB, CII = 7.4, MC = 0.95, WC = 0.97, EME = 7.11 and EMEE = 0.03 | Enhances fine details and reduce excessive enhancement. | Evaluation confined to MR images. | NA |
| Siracusano et al. 2020 [14] | PACE | ENT = 7.69 and CII = 1.31 | Preserves the image details which enhancing contrast and reduced brightness inhomogeneities. | Limited validation on non-COVID-19 datasets and other imaging modalities. | NA |
| Acharya et al. 2021 [15] | Genetic algorithm-based on histogram equalisation | *Entropy = 4.6504, PSNR = 25.0676 dB, SSIM = 0.9176, FSIM = 0.99948, AMBE = 6.8342 and NIQE = 5.1130* | Fully adaptive with automatic parameter selection via GA and effective brightness preservation and contrast enhancement. | Computational complexity and validation limited to a small dataset. | Windows 7, MATLAB 2018 |
| Rawat et al. 2021 [16] | CVMIDNet | PSNR = 37.2010dB and SSIM = 0.9227 | Shows robustness across varying noise levels. | Limited modality testing, future noise types unexplored, and requires more computational resources due to complex-valued operations. | Intel Core i7-8750H (2.20 GHz, 16 GB RAM) with NVIDIA GeForce RTX 2060 GPU. |
| Cao et al. 2021 [17] | Detail-richest-channel based enhancement | Handheld Device: PSNR = 20.45 dB, SSIM = 0.89, NIQE = 2.89 and PIQE = 19.41<br><br>High-End Device: PSNR = 29.64 dB, SSIM = 0.97, NIQE = 2.87 and PIQE = 21.60 | Adaptable to diverse image degradation scenario. | Rquires channel-specific processing for different image types (e.g., retinal vs. underwater). | NA |
| Kumar et al. 2021 [18] | TCDHE-SD, DWT-SVF, SF, IDWT based fusion | COVID-19 CT dataset: SSIM = 0.9432, FSIM = 0.9600, PSNR = 25.569dB, EPI = 0.6321, Entropy = | Effective brightness preservation and edge enhancement and robust against issues of over- or under-enhancement. | Limited validation on non-greyscale images. | NA |
| | | 7.5943, AMBE = 4.3753 and GMSD = 0.0405<br><br>X-ray image dataset: SSIM = 0.8833, FSIM = 0.9488, PSNR = 25.544dB, EPI = 0.7388 and Entropy = 7.0928, AMBE = 6.1079 and GMSD = 0.0596 | | | |
| Voronin et al. 2021 [19] | 3-D block-rooting scheme | NYU dataset: EME = 43.85, AME = 18.59, EMEE =80.93, SDME = 61.25, Visibility = 0.62, TDME = 0.15, BIQI = 59.76, BRISQUE = 9.67, ILNIQE = 22.45, NIQE = 3.22 and GIQEM = 18.66<br><br>FastMRI dataset: EME = 44.38, AME = 20.33, EMEE =46.12, SDME = 62.91, Visibility = 0.61, TDME = 0.31, BIQI = 47.82, BRISQUE = 11.53, ILNIQE = 14.23, NIQE = 2.59 and GIQEM = 14.23<br><br>ChestX-ray dataset: EME = 31.88, AME = 20.78, EMEE =46.12, SDME = 71.52, Visibility = 0.54, TDME | Adaptability to different block sizes and datasets. | Residual noise in large uniform regions. | NA |
| | | = 0.26, BIQI = 55.88, BRISQUE = 14.11, ILNIQE = 25.23, NIQE = 2.78 and GIQEM = 10.15 | | | |
| Jalab et al. 2021 [20] | Fractional calculus-based | Brain MRI:<br>Brisque = 48.5495, Niqe = 4.8101, Histogram flatness = 0.5677 and Histogram spread = 0.0039<br><br>Lung CT:<br>Brisque = 38.9895, Niqe = 2.5339, Histogram flatness = 0.6190 and Histogram spread = 0.0127<br><br>Kidney MRI:<br>Brisque = 28.6598, Niqe = 18.8716, Histogram flatness = 0.8635 and Histogram spread = 0.2485 | Scalable across various datasets. | Slightly limited in handling extremely complex brain MRI images. | MATLAB 2019b |
| Kumar et al. 2022 [21] | Gamma correction and WAHE | VSI = 0.99, CEIQ = 3.23, EBCM = 14.64, NIQE = 3.92 and MEME = 5.00 | Maintains image naturalness and diagnostic relevance. | Edge strength metric (EBCM) slightly underperformed compared to other techniques. | MATLAB R2017a on an i5 laptop (1.19 GHz, 16 GB RAM). |
| Ghosh et al. 2022 [22] | Entropy based intuitionistic fuzzy divergence measure under hyperbolic regularization / HIFDM | MIAS: UQI = 0.8318, SSIM = 0.8357, FQI = 0.8327, IFQI = 0.8477, MAE = 0.1126 and LFI = 0.2859 | Operates minutely in the gray-level dynamic range to highlight small tissue deformities in the breast. | Current scheme does not support automatic detection of abnormal tissue regions, lumps, or masses. | OpenCV and ArrayFire on Python under Ubuntu 20.04 LTS (64-bit, i5 CPU, 16 GB RAM). |
| | | MedPix: UQI = 0.8052, SSIM = 0.8318, FQI = 0.8452, IFQI = 0.8748, MAE = 0.1206 and LFI = 0.3026<br><br>INbreast: UQI =0.8563, SSIM = 0.8298, FQI = 0.8531, IFQI = 0.8836, MAE = 0.1007 and LFI = 0.2642<br><br>DICOM: UQI = 0.8154, SSIM = 0.8431, FQI = 0.8358, IFQI = 0.8426, MAE = 0.1173 and LFI = 0.3046 | | | |
| Huang et al. 2022 [23] | DerivedFuse | Entropy = 7.6314, Contrast improvement index, CII = 1.3095 and average gradient, AG = 8.5074 | Combines classical enhancement methods with deep learning for comprehensive image quality improvement. | Applicability to other medical domains or imaging modalities is untested and the method is computationally intensive. | MATLAB 2019 for image generation, PyTorch 1.5.0 on Intel Xeon E5-2620 (2.10 GHz, 64 GB RAM) with Nvidia Titan Xp GPU |
| Kumar et al. 2022 [24] | Spatial mutual information based | Healthy Brain: NIQMC = 5.337, PCQI = 1.077, RCM = 0.142, MEME = 90.001 and NIQE = 4.821<br><br>Unhealthy Brain: NIQMC = 5.112, PCQI = 1.092, RCM = 0.19, MEME = 104.904 and NIQE = 5.48 | Retains diagnostic information such as tissue structures and boundaries and overcome drawbacks of histogram equalisation. | Computational complexity due to mutual information calculations, the algorithm unable to classify types of brain tumours. | NA |
| | | Multiclass Brain Tumour: NIQMC = 5.363, PCQI = 1.074, RCM = 0.153, MEME = 82.735 and NIQE = 4.812 | | | |
| Kaur et al. 2022 [25] | Hybrid algorithm | PSNR = 27.71dB, FSIM = 0.96, AMBE = 8.38, UIQ = 0.83 and edge content = 9.37 | Retains critical edge and texture details while improving brightness and contrast. | Generalizability to other modalities (e.g., MRI, PET) is untested. | MATLAB 8.5.1 on Windows 10 (2.3 GHz CPU, 6 GB RAM) |
| Liu et al. 2022 [26] | Shannon–Cosine wavelets-based | PSNR = 36.9548dB and SSIM = 0.8297 | Adaptive gain function prevents over-enhancement and noise artifacts. | The algorithm is tailored to linear A/D conversions, making it less effective for other conversion types. | NA |
| Ibrahim et al. 2022 [27] | Fractional partial differential equations (FPDEs) with different types of fractional operators | Brain MRI: Brisque = 40.93, Piqe = 41.13, SSEQ = 66.09 and SAMGVG = 31.04<br><br>For CT Lungs: Brisque = 39.07, Piqe = 41.33, SSEQ = 30.97 and SAMGVG = 159.24 | Superior detail enhancement in low-contrast areas. | Limited effectiveness on complex brain MRI images. | Windows 10 64-bit, Intel Core i7, SSD, 8 GB RAM. |
| Sharif et al. 2022 [28] | Deep Perceptual Enhancement Network | PSNR = 27.61dB and DeltaE = 3.56. | Accelerate CAD application, lightweight and applied in both monochrome and RGB images | Learn from synthesized data samples | AMD Ryzen 3200G (3.6 GHz, 16 GB RAM) with Nvidia GTX 1060 (6 GB). |
| Karim et al. 2022 [29] | FToRE (Fractional Trace Operator with Rényi Entropy) | X-ray Dataset: BRISQUE = 16.4486, PIQE = 21.0140.<br><br>COVID DATABASE: BRISQUE = 36.7163, PIQE = 41.4708. | Empirical tuning of fractional parameters ensures balance between contrast and noise. | Tends to amplify noise in smooth regions. | MATLAB 2021a on Windows 10, Intel i7 (8 GB RAM) with GeForce GTX 950M |
| Abdel-Basset et al. 2022 [30] | T2NS | PSNR = 28.58dB, SNR = 23.60 and SSIM = 0.90 | Handles more complex uncertainties than existing fuzzy method and provide visual and statistical improvements. | Test on other modalities is not involved. | MATLAB R2018a on Windows 10, Intel i7 (2.40 GHz, 8 GB RAM). |
| Navaneetha Krishnan et al. 2022 [31] | Modified optimization approach | *Contrast = 0.9024, PSNR = 55.974dB, weighted PSNR = 38.054dB, homogeneity = 0.9054, SSIM = 0.948 and MSE = 0.0868* | Faster convergence and reduced computational time, along with adaptability to various medical image types. | Testing is constrained by a limited dataset size and the complexity of parameter tuning for MSFO. | MATLAB on Intel Core i5 with 8 GB RAM. |
| Mouzai et al. 2023 [32] | Xray-Net | Cervical spine dataset: AMBE = 0.3954, PSNR = 6.9674 dB, Energy = 6.5782, MSE = 0.2057 and UIQI = 0.0825<br><br>Lumbar spine dataset: AMBE = 0.1771, PSNR = 11.8047 dB, Energy = 6.0556, EME = 10.5984, MSE = 0.0843 and UIQI = 0.1617<br><br>Hand X-rays dataset: AMBE = 0.1726, PSNR = 12.9993 dB, Energy = 6.1806, EME = 6.0479, MSE = 0.0619 and UIQI = 0.2530 | Fully adaptive and self-supervised; no manual adjustments required. | Lacking integration with advanced deep learning for feature-level adjustments. | TensorFlow 2.x, Keras API, Google Colab Pro, Tesla K80 GPU (12 GB), Python. |
| Wu et al. 2023 [33] | Adaptive CLAHE and nonlocal means denoising | Medium-contrast group: Entropy = 7.17, PSNR = 16.02, SSIM = 0.88 and NIQE = 14.71 | Provides optimal trade-off between brightness, contrast, and noise reduction. | Limited to static images; not real-time capable. | NA |
| | | Overexposure group: Entropy = 7.04, PSNR = 23.11, SSIM = 0.90 and NIQE = 15.98<br><br>Small-blood-vessel group: Entropy = 7.25, PSNR = 20.40, SSIM = 0.85 and NIQE = 15.20<br><br>Dense-blood-vessel group: Entropy = 7.09, PSNR = 17.06, SSIM = 0.87 and NIQE = 15.26<br><br>Low-brightness and low-contrast group: Entropy = 6.64, PSNR = 16.80, SSIM = 0.83 and NIQE = 15.31 | | | |
| Ben-Loghfyry et al. 2023 [34] | Regularized Perona–Malik with the Caputo time-fractional order derivative | PSNR = 29.18dB and SSIM = 0.855 | Effective handling of high noise levels and preserves features. | Computational complexity due to the fractional derivative and adaptive numerical schemes; limited to small dataset. | Matlab 2018 on a 3 GHz, 8 GB RAM computer. |
| Sule et al. 2023 [35] | Two-stage histogram equalization enhancement scheme | *DRIVE: PSNR = 42.54203dB, SSIM = 0.92483, MSE = 8.38268 and Euclidean distance = 0.04259*<br><br>*STARE: PSNR = 45.72346dB, SSIM = 0.95928, MSE =* | Balanced global and local enhancements with minimized artifacts. | Computationally intensive due to multi-stage processing and parameter optimization. | MacBook Pro with 2.9 GHz Intel Core i7, 10 GB DDR3 RAM, Intel HD Graphics 4000 (1536 MB), and 148.5 TB shared HDD. Runs macOS with Python 3.7, Scikit-image 0.14.1, OpenCV, and NumPy. |
| | | *4.84619 and Euclidean distance = 0.07731*<br><br>*DIARETDB1: PSNR = 46.90251dB, SSIM = 0.95916, MSE = 3.22932 and Euclidean distance = 0.03947*<br><br>*HRF: PSNR = 48.33635dB, SSIM = 0.96524, MSE = 1.9296 and Euclidean distance = 0.02612* | | | |
| Rao et al. 2023 [36] | DT-CWT and adaptive morphology | PSNR = 27.78dB, entropy = 7.15, CII = 1.67, EME = 17.52, WC = 0.32 and MC = 0.42 | Combines multiscale and adaptive techniques for robust enhancement. | Tailored for CT images, limiting generalizability to other modalities. | Intel Core i5 CPU with 8 GB RAM. |
| Okuwobi et al. 2023 [37] | LTF-NSI | X-ray:<br>EME = 37.02, PSNR = 35.9dB, SSIM = 0.86, p = 0.97, MSE = 21.03, AMBE = 1.55 and SNR = 23.28<br><br>CT:<br>EME = 40.15, PSNR = 38.77dB, SSIM = 0.88, p = 0.98, MSE = 20.08, AMBE = 0.91 and SNR = 25.61 | Robust across multiple modalities | Computational complexity due to optimization. | MATLAB R2013a on Intel Core i5-4200U (1.60 GHz, 8 GB RAM). |
| | | Optical Coherence Tomography Angiography (OCTA): EME = 40.01, PSNR = 38.01dB, SSIM = 0.87, p = 0.98, MSE = 22.15, AMBE = 1.02 and SNR = 26.73<br><br>Fluorescein Angiography (FA): EME = 42.14, PSNR = 40.15dB, SSIM = 0.89, p = 0.98, MSE = 19.52, AMBE = 0.88 and SNR = 30.55 | | | |
| Yu et al. 2023 [38] | FS-GAN | Entropy = 6.785, AvG = 7.332, Brisque = 0.484, NIQE = 28.107 and PIQE = 1.774 | Effective unpaired learning with strong structural preservation and novel application of fuzzy theory in GANs. | High computational costs from GAN complexity and unexamined applicability to other modalities. | Ubuntu 18.04 with Nvidia GeForce RTX 3090. |
| Jiang et al. 2023 [39] | Group theoretic particle swarm optimization (GT-PSO) | Fitness scores = 11.885 | Superior performance in optimizing multi-modal and non-linear intensity transformations. | High computational complexity due to group theoretic operations and no guarantee of reaching global optima inherent to metaheuristic methods. | NA |
| Pashaei et al. 2023 [40] | Arithmetic Optimization Algorithm (AOA) | *SSIM = 0.84406, SE = 6.31122, PSNR = 22.67356 dB, AMBE = 0.03984, NIQE = 3.3985 and QI =0.72816* | Dynamically adjusts parameters, ensuring consistent enhancement across diverse image sets. | Computational overhead due to iterative optimization and Gaussian mutation and the performance dependent on parameter | MATLAB R2019a on Intel Core i5 (2.4 GHz, 8 GB RAM). |
| | | | | initialization and fitness function design. | |
| Zhong et al. 2023 [41] | MAGAN | PSNR = 15.31 dB, SSIM = 0.793, Entropy = 6.796, AvG = 7.212, Brisque = 0.491, NIQE = 30.177 and PIQE = 1.829 | Superior performance in downstream segmentation tasks. | Misidentifies large artifacts as nerve fibers in some cases and suffers from structural degradation in areas with very unclear features. | Ubuntu 18.04, Intel Xeon Gold 633 (2.00 GHz, 48 GB RAM), with Nvidia GeForce RTX 3090. |
| Mousania et al. 2023 [42] | Optimal new histogram equalization technique / BPDF-min CE | Mammograms: PSNR = 37.05, EME = 9.73, MSE = 41.13, minimal AMBE = 0.02 and SSIM = 0.97<br><br>Carotid artery: PSNR = 34.12, EME = 12.91, MSE = 46.88, minimal AMBE = 0.05 and SSIM = 0.97<br><br>Focal liver lesions: PSNR = 35.74, EME = 16.69, MSE = 43.09, minimal AMBE = 0.07 and SSIM = 0.98<br><br>Brain CT Scan: PSNR = 34.96, EME = 16.31, MSE = 46.65, minimal AMBE = 0.01 and SSIM = 0.98<br><br>Brain MRI: PSNR = 37.18, EME = 9.05, MSE = 39.73, minimal | Less computational complexity, preserves brightness and enhancement contrast across different types of images. | Computational time slightly higher due to iterative optimization. | NA |
| | | AMBE = 0.01 and SSIM = 0.98 | | | |
| Trung 2023 [43] | Fuzzy logic Clustering-based | *Std = 0.078344 and Sharp index = 0.072311* | Localized enhancement improves dark object visibility without over-enhancing bright areas and robust against varying brightness levels across different image regions. | Computational complexity due to the clustering and iterative enhancement process and dependence on parameter settings for clustering and enhancement bounds. | NA |
| Jiang et al. 2024 [44] | ARM-Net v2 | Spatial resolution of 128 x 128:<br>PSNR = 36.8271dB, SSIM = 0.9568 and LPIPS = 0.0529 | Robust handling of Rician noise and low computational cost. | Adaptation for other modalities may limited. | 3 NVIDIA GeForce RTX 2080 Ti GPUs in parallel on CentOS Linux |
| Guo et al. 2024 [45] | Multi-Degradation-Adaptive-Net | EyeQ 'Good':<br>PSNR = 35.52 dB and SSIM = 0.9692<br>EyeQ 'Usable':<br>WFQA = 1.2102 and FIQA = 0.2635<br>EyeQ 'Reject':<br>WFQA = 0.3259 and FIQA = 0.0305<br><br>DRIVE:<br>PSNR = 28.76 dB and SSIM = 0.7431<br><br>REGUGE:<br>PSNR = 26.29 dB and SSIM = 0.8873 | Robustness to unknown degradation levels and types via contrastive learning. | High computational cost due to dynamic filter generation and representation learning. | PyTorch and a single NVIDIA RTX A4500 GPU. |
| Acharya et al. 2024 [46] | DSOTAGC | Entropy = 6.01, PSNR = 22.557 dB, AMBE = 17.178 and SSIM = 0.921 | Adaptive for diverse image types due to optimized parameters. | Did not apply on RGB images. | MATLAB R2018a |
| Xu et al. 2024 [47] | Siamese-based structure, GAN | Entropy = 6.6951, AvG = 8.5481, NIQE = 3.9778 and PIQE = 5.3141 | Addresses structural preservation and robustness to noise. | Requires computational resources due to GAN-based architecture. | - |
| Chandra et al. 2024 [48] | Modified Type II fuzzy set | Ultrasound images:<br>AMBE = 0.57, entropy = 5.08, PSNR = 50.35, SSIM = 0.99, PL measure = 336.38 and REC = 0.97<br><br>MRI images:<br>AMBE = 1.11, entropy = 5.66, PSNR = 45.52, SSIM = 0.99, PL measure = 134.14 and REC = 0.99<br><br>X-ray images:<br>AMBE = 2.34, entropy = 7.44, PSNR = 39.35, SSIM = 0.99, PL measure = 73.84 and REC = 0.99 | Improved contrast enhancement with minimal over-brightness. | Performance depends on parameter tuning (α). | MATLAB (2016) |
| Cap et al. 2025 [49] | LaMEGAN | LaSSIM = 0.936; MDOS-O = 4.05, NIMA = 4.05, PaQ-2-PiQ = 74.91, DBCNN = 58.03, MUSIQ = 56.36, HyperIQA = 53.39, MDOS-Q = 3.67, NIQE = 4.45 and BRISQUE = 21.96 | A robust metric for non-reference structural evaluation is introduced. | Occasional production of bold red areas results in unnatural and unrealistic color distribution and LaSSIM scores are only valid for relative comparisons, primarily based on throat images with limited validation in diverse medical datasets. | NVIDIA V100 GPU with 16GB |

** *Italicized* text signifies that the average results were self-calculated for consistency, as the original paper lacked an average score.

** NA indicates not available.

Table 3.5 Analysis of reference-based IQA metrics in reviewed studies

| Metrics | Concept | Equation | Indications | Reference |
|---|---|---|---|---|
| Mean-Squared Error (MSE) | Measures the average squared difference between the original image and the processed image. | $MSE = \frac{1}{MN}\sum_{i=1}^{M}\sum_{j=1}^{N}\left(I(i,j) - K(i,j)\right)^2$<br>$I(i,j) = Original\ image\ pixel\ value$<br>$Y(i,j) = Processed\ image\ pixel\ value$<br>$M, N = Image's\ dimensions$ | 0 | [11], [13], [31], [32], [35], [42] |
| Peak Signal-to-Noise Ratio (PSNR) | Measures the ratio between the maximum possible pixel value and the noise present in the image. Higher values indicate better quality. | $PSNR = 10\log_{10}\left(\frac{(MAX^2)}{MSE}\right)$<br>$MAX_I = Maximum\ possible\ pixel\ value$ | 1 | [13], [15], [16], [17], [18], [26], [28], [30], [31], [32], [33], [34], [35], [36], [40], [42], [44], [45], [46], [48] |
| Signal to Noise Ratio (SNR) | A measure of the ratio of the signal power to the noise power in an image. | $SNR = \frac{\mu^2}{\sigma^2}$<br>$\mu = The\ mean$<br>$\sigma^2 = The\ variance\ of\ the\ image$ | 1 | [30] |
| Weighted PSNR | A variant of PSNR that gives more weight to certain regions of the image. | $WPSNR = 10\log_{10}\left(\frac{(MAX)^2}{MSE \times Noise\ visibility\ function}\right)$ | 1 | [31] |
| Structural Similarity Index (SSIM) [90], [91] | Measures perceptual similarity between two images, considering luminance, contrast, and structure.<br><br>0-1 | $SSIM(x,y) = \frac{(2\mu_x\mu_y + C_1)(2\sigma_{xy} + C_2)}{(\mu_x^2 + \mu_y^2 + C_1)(\sigma_x^2 + \sigma_y^2 + C_2)}$<br>$\mu_{x,y} = Mean\ intensity\ of\ images\ x\ and\ y$<br>$\sigma_x^2, \sigma_y^2 = Variance\ of\ images\ x\ and\ y$<br>$\sigma_{x,y} = Covariance\ between\ x\ and\ y$<br>$C_{1,2} = Small\ constants\ to\ avoid\ zero\ division$ | 1 | [11], [15], [16], [17], [18], [22], [26], [30], [31], [33], [34], [35], [37], [40], [42], [44], [45], [46], [48] |
| Feature Similarity | Measures similarity between two images based on low-level features | $FSIM = \sum_{i}\sum_{j} w_1(i,j) \cdot \lvert G_1(i,j) - G_2(i,j)\rvert + w_2(i,j) \cdot \lvert P_1(i,j) - P_2(i,j)\rvert$ | 1 | [15], [18], [35] |
| Index Measurement (FSIM) [92] | like gradient magnitude and phase congruency. 0-1 | $G_1, G_2 = Gradient\ magnitudes$<br>$P_1, P_2 = Phase\ congruencies$ | | |
| Universal Quality Index (UQI) | Measures the perceptual quality by considering correlation, luminance, and contrast between the reference and distorted images. | $$UQI = \frac{2\sigma_{xy} + c_1}{\sigma_x^2 + \sigma_y^2 + c_1} \cdot \frac{2\mu_x\mu_y + c_2}{\mu_x^2 + \mu_y^2 + c_2}$$<br>$\mu_x, \mu_y = The\ mean\ intensities$<br>$\sigma_x, \sigma_y = Standard\ deviations$<br>$\sigma_{xy} = The\ cross-correlation$ | 1 | [22], [32] |
| Edge Preservation Index (EPI) [93] | Measures how well edges are preserved in an enhanced or processed image. | $$EPI = \frac{\sum \lvert G_Y \rvert}{\sum \lvert G_X \rvert}$$<br>$G_X = Gradient\ magnitudes\ of\ the\ original\ images$<br>$G_Y = Gradient\ magnitudes\ of\ the\ processed\ images$ | 1 | [11], [18] |
| Absolute Mean Brightness Error (AMBE) [91] | Measures brightness difference between the original and enhanced image. | $$AMBE = \lvert \mu_x - \mu_y \rvert$$<br>$\mu_x = Mean\ brightness\ of\ input\ image$<br>$\mu_y = Mean\ brightness\ of\ enhanced\ image$ | 0 | [15], [18], [32], [37], [40], [42], [46], [48] |
| Gradient Magnitude Similarity Deviation (GMSD) [94] | Measures the deviation in gradient magnitude between an image and its reference, indicating image quality. | $$GMSD = \frac{1}{MN}\sum_{i=1}^{M}\sum_{j=1}^{N}(\lvert\nabla I(i,j)\rvert - \lvert\nabla K(i,j)\rvert)^2$$<br>$\lvert\nabla I(i,j)\rvert)\ and\ \lvert\nabla K(i,j)\rvert$<br>$= The\ gradient\ magnitudes\ of\ the\ reference\ and\ distorted\ images$ | 0 | [18] |
| Visual Saliency Induced Index (VSI) [95] | Measures the perceptual quality of an image by considering visual saliency and information content. | $$VSI = \frac{\sum_{i=1}^{N} \lvert S_i, I_i \rvert}{\sum_{i=1}^{N} \lvert S_i, K_i \rvert}$$<br>$S_i = Saliency\ map$<br>$I_i = Pixel\ intensity\ of\ input\ image$<br>$K_i = Pixel\ intensity\ of\ the\ reference\ image$ | 1 | [21] |
| Relative Enhancement in Contrast (REC) | Measures the improvement in contrast between the processed and original image. This can be done by adjusting the darkness and brightness of objects. | $$REC = \frac{C_Y}{C_X}$$<br>$C_X = Contrast\ levels\ of\ the\ original\ images$<br>$C_Y = Contrast\ levels\ of\ the\ processed\ images$ | 1 | [11], [48] |
| Contrast Improvement Index (CII) [96] | Measures of how much contrast has been enhanced in an image. This can be done by adjusting the darkness and brightness of objects. | $$CII = \frac{C_e}{C_X}$$<br>$C_e = Contrast\ levels\ of\ the\ processed\ images$<br>$C_X = Contrast\ levels\ of\ the\ original\ images$ | 1 | [13], [14], [23], [36] |
| Homogeneity | Measures the uniformity of the image pixel values. | $H = \frac{1}{MN}\sum_{i=1}^{M}\sum_{j=1}^{N}\lvert I(i,j) - \mu\rvert$ | 1 | [31] |
| Mean absolute error (MAE) | Measures the average of the absolute differences between the reference and the distorted image. | $MAE = \frac{1}{MN}\sum_{i=1}^{M}\sum_{j=1}^{N}\lvert I(i,j) - K(i,j)\rvert$<br>$I(i,j)\ and\ K(i,j)) = The\ pixel\ values\ of\ the\ reference\ and\ distorted\ images$ | 0 | [22] |
| Linear Fuzzy Index (LFI) | Measures image quality based on fuzzy logic, considering the fuzziness in pixel intensities. | $LFI = \sum_{i=1}^{M}\sum_{j=1}^{N}\lvert I(i,j) - K(i,j)\rvert$ | 0 | [22] |
| Fuzzy Quality Index (FQI) [97] | A fuzzy logic-based method for assessing the quality of images by comparing the reference and distorted images. | $FQI = \sum_{i=1}^{M}\sum_{j=1}^{N}\left(\frac{\lvert I(i,j) - K(i,j)\rvert}{1 + \lvert I(i,j) - K(i,j)\rvert}\right)$ | 1 | [22] |
| Intuitionistic Fuzzy Quality Index (IFQI) [98] | Similar to FQI, but it considers the uncertainty in the image content using intuitionistic fuzzy sets. | $IFQI = \sum_{i=1}^{M}\sum_{j=1}^{N}\left(\frac{\lvert I(i,j) - K(i,j)\rvert}{1 + \sqrt{\lvert I(i,j) - K(i,j)\rvert^2 + \lvert I(i,j) - K(i,j)\rvert}}\right)$ | 1 | [22] |
| Quality Index (QI) | Measures image quality based on loss of correlation, luminance distortion, and contrast distortion.<br><br>0 - 1 | $QI = \frac{4\sigma_{xy}\mu_x\mu_y(\mu_x^2 + \mu_y^2)}{(\sigma_x^2 + \sigma_y^2)}$<br>$\mu_{x,y} = Mean\ intensity\ of\ reference\ and\ test\ images$<br>$\sigma_x^2\ ,\ \sigma_y^2 = Variance\ of\ reference\ and\ test\ images$<br>$\sigma_{x,y} = Covariance\ between\ reference\ and\ test\ images$ | 1 | [40] |
| Relative Contrast Measure (RCM) [99] | A metric that evaluates the relative contrast change between an original and an enhanced image. It assesses how much contrast improvement or degradation has occurred due to an enhancement process. | $RCM = \sum_{m}\sum_{n}\Delta G(m,n)W(m,n)$<br>$G(m,n) = Relative\ gradient\ change\ between\ refernece\ and\ enhanced\ image$<br>$W(m,n) = Weighting\ function\ that\ emphasizes\ significant\ edge\ regions$ | 1 | [24] |
| Edge Content (EC) | Evaluates the gradient magnitude of contrast variations, it quantifies how much contrast has improved in the processed image relative to the original. | $EC = \frac{1}{m \times n}\sum_{x=1}^{m}\sum_{y=1}^{n}\sqrt{g_x^2(x,y) + g_y^2(x,y)}$<br>$m \times n = Height\ and\ width\ of\ image$<br>$x\ and\ y = Pixel\ coordinates$<br>$g_{x,y}^2(x,y) = Horizontal\ and\ vertical\ gradient$ | 1 | [25] |
| Learned Perceptual Image Patch Similarity (LPIPS) [50] | A deep learning-based metric for measuring perceptual similarity between image patches. | $LPIPS = \frac{1}{N}\sum_{i=1}^{N}(\text{feature distance between patches})$ | 0 | [44] |
| PL Measure | The ratio of Peak Signal-to-Noise Ratio (PSNR) to the Linear Fuzziness Index (LFI). It quantifies the amount of fuzziness present in an enhanced image. | $PL = \frac{PSNR}{c}$<br>$c = Linear\ fuzziness\ index$ | 1 | [48] |

** 1 = A higher metric value indicates better image quality.

** 0 = A lower metric value indicates better image quality.

Table 3.6 Analysis of non-reference-based IQA metrics in reviewed studies

| Metrics | Concept | Equation | Indications | Reference |
|---|---|---|---|---|
| Fitness Function / Scores | A fitness function that optimizes image enhancement by maximizing edge intensity, edge pixel count, and entropy using weighted correlation. | Depends on the specific application | 1 | [12], [39] |
| Histogram flatness | Measures how evenly the histogram of an image is distributed. A flat histogram indicates a uniform distribution of pixel intensities. | $HF = \frac{(\prod_{i=1}^{L} h(i))^{\frac{1}{L}}}{\frac{1}{L}\sum_{i=1}^{L} h(i)}$<br>$h(i) = The\ histogram\ count\ at\ intensity\ level\ i$<br>$L = \ The\ total\ number\ of\ intensity\ levels$<br>$\prod_{i=1}^{L} h(i)$ = The product of all histogram counts<br>$\sum_{i=1}^{L} h(i)$ = The sum of all histogram counts | 1 | [20] |
| Histogram spread | Measures the spread (or dispersion) of the histogram values, reflecting the contrast of the image. | $HS = \frac{Q_3 - Q_1}{R}$<br>$Q_1 = 25th\ percentile\ of\ the\ histogram\ bin\ positions$<br>$Q_3 = 75th\ percentile\ of\ the\ histogram\ bin\ positions$<br>$R = Possible\ range\ of\ image\ pixel\ values$ | 1 | [20] |
| Edge-Based Contrast Measure (EBCM) [100] | Measures the contrast based on edge strength and the number of edges in an image. | $EBCM = \frac{\sum_i \lvert \nabla I(i,j) \rvert}{\sum_{i,j} I(i,j)}$<br>$\nabla I(i,j) = The\ gradient\ (edge)\ of\ the\ image$ | 1 | [21] |
| Entropy (ENT), Shannon Entropy (SE) / Information entropy [101] | Measure of the randomness or unpredictability of pixel intensities in an image in terms of texture or detail. | $H(x) = -\sum_{i=1}^{N} p(i) \log_2 p(i)$<br>$p(i) = Probability\ of\ the\ pixel\ intensity\ i\ occurring$ | 1 | [11], [12], [13], [14], [15], [18], [23], [35], [36], [38], [41], [46], [47], [48] |
| Number of Edges | This refers to the number of significant transitions (edges) in an image. It is often used to | The number of edges is computed based on edge detection algorithms like Canny or Sobel filters. | 1 | [12] |
| | measure the sharpness and detail of an image. | | | |
| Sharpness Index [102] | Sharpness index measures the level of edge clarity or fine detail in an image. It is commonly used to evaluate how crisp the image appears. | It is precisely calculated using six Discrete Fourier Transforms (DFTs). $$SI(u) = -\log_{10} \Phi\left(\frac{\mu - \mathrm{TV}(u)}{\sigma}\right)$$ $SI(u) = Sharpness\ index$<br>$\Phi(\cdot) = CDF\ of\ standard\ normal\ distribution$<br>$u = Mean\ TV\ value$<br>$TV(u) = Total\ Variation\ of\ image$<br>$\sigma = Standard\ deviation\ of\ TV\ values$ | 1 | [12], [21], [43] |
| Simplified Sharpness of a Numerical Image [103] | The sharpness of a numerical image can be understood probabilistically, as it exhibits unexpectedly low total variation compared to related random-phase fields. | $$S = \frac{\sum_{i,j} \lvert I(i,j) - I(i-1,j)\rvert + \lvert I(i,j) - I(i,j-1)\rvert}{N}$$ $I(i,j) = Pixel\ value; N = The\ number\ of\ pxiels$ | 1 | [12] |
| No-reference Image Quality metric for Contrast Distortion (NIQMC) [104] | NIQMC is a no-reference image quality metric that evaluates contrast-altered images by maximizing information entropy, prioritizing local details, and comparing unpredictable components to the full image to estimate visual quality. | $$NIQMC = \frac{L' + \alpha G'}{1 + \alpha}$$ $L = Local\ entropy - based\ quality\ measurement$<br>$G' = Global\ histogram\ based\ quality\ assessment$<br>$\alpha = Weighting\ factor$ | 1 | [24] |
| Contrast Enhanced Image Quality Index (CEIQ) [105] | Evaluates the contrast enhancement quality of an image. It quantifies the enhancement applied to the contrast while maintaining natural features. | $$CEIQ = f(S_{ge}, E_g, E_e, E_{ge}, E_{eg})$$ $S_{ge} = SSIM$<br>$E_g, E_e = Entropies\ of\ the\ grayscale\ and\ enhanced\ images$<br>$E_{ge}, E_{eg}$<br>$= Cross - entropy\ values\ computed\ between\ the\ histograms\ of\ two\ images$ | 1 | [12], [21] |
| Contrast (C) | The difference in luminance or color makes an object distinguishable. In images, it measures the contrast between the darkest and lightest points. | $$C = \frac{Maximum\ pixel\ value - Minimum\ pixel\ value}{Maximum\ pixel\ value + Minimum\ pixel\ value}$$ | 1 | [13], [31] |
| Michelson Contrast [106] | Measure used to quantify the contrast of periodic or | $$MC = \frac{I_{max} - I_{min}}{I_{max} + I_{min}}$$ | 1 | [13], [36] |
| | sinusoidal patterns, commonly applied to images with periodic textures. | $I_{max}, I_{min} = Maximum\ and\ minimum\ pixel\ intensities$ | | |
| Weber Contrast [106] | **Measures the contrast between a target and its surrounding background.** | $$\boldsymbol{C_{Weber} = \frac{I_{target} - I_{background}}{I_{background}}}$$ | **1** | [13], [36] |
| Measure of Enhancement (EME) / Measure of Improvement / Modified Measure of Enhancement (MEME) [107], [108] | Evaluates the effectiveness of image enhancement by measuring changes in contrast or other quality aspects before and after enhancement. | $$\text{EME}_{m_1 m_2}(\Psi) = \frac{1}{m_1 m_2}\sum_{p=1}^{m_1}\sum_{q=1}^{m_2} 20\ln\left(\frac{J^v_{max:p,q}}{J^v_{min:p,q}}\right)$$ $m_1 m_2 = The\ number\ of\ segmentaed\ blocks\ in\ the\ image$ $J_{max}\ and\ J_{min} = The\ maximum\ and\ minimum\ intensity\ values$ | 1 | [13], [14], [19], [21], [24], [32], [36], [37], [42] |
| Measure of Enhancement by Entropy (EMEE) [107] | Evaluates how well the enhancement process has increased the image's entropy, which correlates to more detailed or informative content. | $$\text{EMEE}_{m_1 m_2}(\Psi) = \frac{1}{m_1 m_2}\sum_{p=1}^{m_1}\sum_{q=1}^{m_2} \alpha\left(\frac{J^v_{max:p,q}}{J^v_{min:p,q}}\right)^{\propto} \cdot \ln\left(\frac{J^v_{max:p,q}}{J^v_{min:p,q}}\right)$$ | 1 | [13], [19] |
| Visibility [109] | The Michelson Visibility Operator is a contrast measurement method used to quantify the strength of interference fringes in an image. It is applied in infrared image enhancement and target detection to improve the visibility of dim objects. | $$Visibility_{m_1 m_2}(\Psi) = \sum_{p=1}^{m_1}\sum_{q=1}^{m_2} \frac{J^v_{max:p,q} - J^v_{min:p,q}}{J^v_{max:p,q} + J^v_{min:p,q}}$$ | 1 | [19] |
| AME [108] | AME quantifies contrast based on Michelson's Contrast Law in a logarithmic domain. | $$AME_{m_1 m_2}(\Psi) = -\frac{1}{m_1 m_2}\sum_{p=1}^{m_1}\sum_{q=1}^{m_2} 20\ln\left(\frac{J^v_{max:p,q} - J^v_{min:p,q}}{J^v_{max:p,q} + J^v_{min:p,q}}\right)$$ | 1 | [19] |
| Second Derivative based Measure (SDME) [110] | It evaluates the rate of change in pixel intensity variations while also accounting for the center pixel value along with | $$SDME_{m_1 m_2}(\Psi) = -\frac{1}{m_1 m_2}\sum_{p=1}^{m_1}\sum_{q=1}^{m_2} 20\ln\left(\frac{J^v_{max:p,q} - 2J^v_{center:p,q} + J^v_{min:p,q}}{J^v_{max:p,q} + 2J^v_{center:p,q} + J^v_{min:p,q}}\right)$$ | 1 | [19] |
| | the local maximum and minimum values. | | | |
| Transform domain measure of enhancement (TDME) [111] | Analyzing changes in high-frequency components in the Discrete Cosine Transform (DCT) domain. | $$TDME = \frac{\sum_{i=1}^{M}\sum_{j=1}^{N}\lvert C_H(i,j)\rvert}{\sum_{i=1}^{M}\sum_{j=1}^{N}\lvert C(i,j)\rvert}$$ $C_H(i,j) = High\ frequency\ DCT\ coefficient\ at\ position\ i\ and\ j$ <br> $C(i,j) = DCT\ coefficient\ at\ position\ i\ and\ j$ <br> $M, N = Dimensions\ of\ the\ DCT\ coefficient\ matrix$ | 1 | [19] |
| Average Gradient (AG) | Measures the average gradient magnitude of an image, indicating its sharpness and texture. | $$AG = \frac{1}{MN}\sum_{i=1}^{M}\sum_{j=1}^{N}\lvert\nabla I(i,j)\rvert$$ $MN = Size\ of\ image$ <br> $\nabla I(i,j) = Gradient\ in\ vertical\ and\ horizontal$ | 1 | [23], [38], [41], [47] |
| Patch-based Contrast Quality Index (PCQI) [112] | Measures the contrast quality in local patches of the image. | $$PCCQI = C' \cdot S' \cdot M'$$ $C' \cdot S' \cdot M' = Contrast\ variation, structural\ similarity\ and\ mean\ intensity\ difference$ | 1 | [24] |
| Spatial–Spectral Entropy-based Quality (SSEQ) index [113] | Measures the entropy of the image in both spatial and spectral domains. | $$SSEQ = -\sum_{i=1}^{M}\sum_{j=1}^{N} p(i,j)\log\big(p(i,j)\big)$$ $p(i,j)) = The\ probability\ distribution\ of\ pixel\ values$ | 0 | [27] |
| The Blind Image Sharpness Assessment Based on Maximum Gradient and Variability of Gradients (SAMGVG) [114] | A sharpness measure based on the maximum gradient and its variability in the image. | $$SAMGVG = \max(\nabla I) + \mathrm{Variance}(\nabla I)$$ | 1 | [27] |
| DeltaE [115] | Measures the perceptual difference between two images in terms of color space. | $$\Delta E = \sqrt{(L^* - L_0)^2 + (a^* - a_0)^2 + (b^* - b_0)^2}$$ | 0 | [28] |
| Perception-based Image Quality Evaluator (PIQE) [116] | Evaluates the image quality by considering various perceptual features, such as contrast, sharpness, and blur. | $$PIQE = \frac{\sum_{i=1}^{N_A} S_i + C}{N_A + C}$$ $S_i = Distortion\ score\ for\ block\ i$ <br> $N_A = Number\ of\ active\ blocks\ in\ the\ image$ <br> $C = Small\ constant\ to\ prevent\ numerical\ instability$ | 0 | [17], [29], [38], [41], [47] |
| Blind/Reference-less Image Spatial Quality Evaluator (BRISQUE) [117] | Evaluates quality based on spatial domain features. | $$I_{\text{MSCN}}(i,j) = \frac{I(i,j) - \mu(i,j)}{\sigma(i,j) + C}$$ $I(i,j)) = \textit{The pixel intensity}$ <br> $\mu(i,j) = \textit{The local mean}$ <br> $\sigma(i,j)) = \textit{The local variance}$ <br> $C = \textit{Small constant to prevent division by zero}$ | 0 | [19], [20], [27], [29], [38], [41], [49] |
| Natural Image Quality Evaluator (NIQE) [118] | Evaluates the quality of an image by comparing its statistical features with a natural image database. | $$NIQE = D\big((\mu_n, \Sigma_n), (\mu_d, \Sigma_d)\big)$$ $(\mu_n, \Sigma_n) = \textit{The mean and covariance of the natural image model.}$ <br> $(\mu_d, \Sigma_d) = \textit{The mean and covariance of the distorted image.}$ <br> $D(.,.) = \textit{Mahalanobis distance or a similar statistical measure.}$ | 0 | [12], [15], [17], [19], [20], [21], [24], [27], [33], [38], [40], [41], [47], [49] |
| Energy | **Measures the energy of the image, which can indicate the sharpness or clarity of the image.** | $$E = \sum_{i=1}^{M}\sum_{j=1}^{N} I(i,j)^2$$ | 1 | [32] |
| Standard deviation | Measures the spread or variation of pixel values in the image. | $$\sigma = \sqrt{\frac{1}{MN}\sum_{i=1}^{M}\sum_{j=1}^{N}(I(i,j) - \mu)^2}$$ | 1 | [43] |
| Euclidean distance [119] | Measures the distance between two image vectors in a multi-dimensional space. | $$d = \sqrt{\sum_{i=1}^{M}\sum_{j=1}^{N}\big(I(i,j) - K(i,j)\big)^2}$$ | 0 | [35] |
| Fundus Image Quality Assessment (FIQA) [120] | Evaluates the quality of fundus images for medical applications. | $FIQA$ = Features of image: sharpness, contrast, and noise | 1 | [45] |
| Weighted FIQA (WFQA) | A weighted version of FIQA that considers the importance of different image features. | $$WFQA = \sum_{i=1}^{N} w_i \cdot f_i$$ $w_i = \textit{The weights}$ <br> $f_i = \textit{The individual features of the image}$ | 1 | [45] |
| Integrated Local Natural Image | It models natural scene statistics (NSS) features using a multivariate Gaussian (MVG) | $$Q = \frac{1}{k}\sum_{i=1}^{k} d_i$$ | 0 | [19] |
| Quality Evaluator (IL-NIQE) [121] | model from pristine images and compares test images against this reference model. | $d_i = Distortion\ level\ of\ patch\ i\ measure\ using\ Bhattacharyya\ distance$ | | |
| Laplacian Structural Similarity Index Measure (LaSSIM) | It evaluates the structural preservation of medical images by applying Laplacian Pyramid (LP) decomposition before computing SSIM. | $LaSSIM_l(I, I_b) = SSIM\left(LP_l(I), LP_l(I_b)\right)$ <br> $LP_l = residual\ signal\ at\ level\ l$ <br> $I\ and\ I_b = Original\ and\ enhanced\ images, respectively$ | 1 | [49] |
| Blind Image Quality Index (BIQI) [122] | Assesses image quality by extracting scene statistics and using them to classify distortion types and predict quality scores. | $BIQI = f(S)$ <br> $S = The\ scene\ statistics\ extracted\ from\ the\ distorted\ image$ <br> $f(\ \ ) = A\ regression\ model\ that\ maps\ extracted\ features$ | 1 | [19] |
| Neural Image Assessment (NIMA) [51] | NIMA evaluates both technical and aesthetic image quality. The model is trained using deep convolutional neural networks (CNNs) to predict human opinion scores. | $Q_{nr} = g(I)$ <br> $Q_{nr} = The\ quality\ score$ <br> $I = The\ input\ image$ <br> $g(\cdot) = A\ function\ that\ predicts\ quality\ based\ on\ learned\ features$ | 1 | [49] |
| From Patches to Pictures (PaQ-2-PiQ) [52] | The PaQ-2-PiQ model predicts perceptual image quality by analyzing local patches and mapping them to a global image quality score using deep learning techniques. It leverages a region-based deep neural network to infer both local patch quality and global picture quality. | $PaQ\text{-}2\text{-}PiQ = f_\theta(I)$ <br> $f_\theta = The\ deep\ neural\ network\ model\ trained\ on\ human - annotated\ quality\ labels$ <br> $I = The\ input\ image\ whose\ quality\ is\ being\ predicted$ | 1 | [49] |
| Deep bilinear convolutional neural network (DBCNN) [53] | DBCNN is a deep learning-based no-reference image quality assessment model that combines two CNNs: <br><br> • A CNN trained on synthetic distortions (e.g., | $DBCNN = f_\theta(I)$ <br> $Q = The\ predicted\ image\ quality\ score$ <br> $f_\theta = The\ deep\ bilinear\ CNN\ model\ trained\ to\ map\ image\ features$ <br> $I = The\ input\ image\ being\ evaluated$ | 1 | [49] |
| | compression artifacts, blur, noise).<br>• A CNN pre-trained for general image classification (e.g., authentic distortions from real-world images). | | | |
| HyperIQA [54] | Employs a self-adaptive hyper network to dynamically generate content-aware quality prediction parameters, enabling improved generalization and alignment with human perception. | $$\phi(x, H(S(x), \gamma)) = q,$$<br>$\phi = Quality\ assessment\ function$<br>$x = Input\ image$<br>$S(x) = \ Extracted\ semantic\ features$<br>$H(S(x), \gamma) = \ The\ hypernetwork\ that\ maps\ the\ extracted\ features$<br>$q = Predicted\ image\ quality\ score$ | 1 | [49] |
| Multi-scale Image Quality (MUSIQ) [55] | It utilizes multi-scale image representation, hash-based 2D spatial embedding, and scale embedding to predict perceptual quality directly from raw images. | $$MUSIQ = g_\theta(I)$$<br>$Q = The\ predicted\ image\ quality\ score$<br>$g_\theta = The\ multi-scale\ Transformer\ model$<br>$I = The\ input\ image\ being\ evaluated$ | 1 | [49] |
| Golden Image Quality Enhancement Measure, (GIQEM) | Measures contrast enhancement using the Golden transform by capturing high-frequency content. | $$GIQEM = \frac{1}{k_1 k_2} 20log\left(\frac{\sum\lvert\breve{G}\rvert}{\lvert\breve{G}_{max}\rvert}\right)$$<br>$k_1 k_2 = Size\ of\ the\ sliding\ block$<br>$\breve{G} = Transformed\ image\ coefficients$<br>$\sum\lvert\breve{G}\rvert = Sum\ of\ the\ magnitude\ of\ high-frequecy\ components$<br>$\lvert\breve{G}_{max}\rvert = Maximum\ magnitude\ of\ trasnformed\ coefficients$ | 1 | [19] |

** 1 = A higher metric value indicates better image quality.

** 0 = A lower metric value indicates better image quality.